\documentclass[11pt]{article}

\usepackage[letterpaper, margin=1in]{geometry}
\usepackage{amsmath,amssymb,amsthm}
\usepackage{mathtools}
\usepackage{graphicx}
\usepackage{hyperref}
\usepackage{url}
\usepackage{xcolor}
\usepackage{enumitem}
\usepackage{booktabs}
\usepackage{microtype}
\usepackage[round,authoryear]{natbib}
\usepackage{mathptmx}

\hypersetup{
  colorlinks=true,
  linkcolor=blue!60!black,
  citecolor=blue!60!black,
  urlcolor=blue!60!black,
  pdftitle={How Much Cache Does Reasoning Need? Depth-Cache Tradeoffs in KV-Compressed Transformers},
  breaklinks=true
}

\newtheorem{theorem}{Theorem}[section]
\newtheorem{lemma}[theorem]{Lemma}
\newtheorem{proposition}[theorem]{Proposition}
\newtheorem{corollary}[theorem]{Corollary}
\newtheorem{conjecture}[theorem]{Conjecture}
\theoremstyle{definition}
\newtheorem{definition}[theorem]{Definition}
\theoremstyle{remark}
\newtheorem{remark}[theorem]{Remark}

\newcommand{\PC}{\mathrm{PC}}

\title{How Much Cache Does Reasoning Need?\\
\large Depth--Cache Tradeoffs in KV-Compressed Transformers}
\author{%
  Xiao Wang  \\ \texttt{FujianAI42@163.com}
}
\date{}

\begin{document}
\maketitle

\begin{abstract}
The key-value (KV) cache is the dominant memory bottleneck during Transformer inference, yet little is known theoretically about how aggressively it can be compressed before multi-step reasoning degrades. We study this question through $k$-hop pointer chasing under a shared KV cache of size $s$, attention dimension $m$, $H$ heads, $p$-bit precision, and a locality-respecting cache controller (a minimal restriction satisfied by all standard KV-compression methods). We give three results.

\emph{(1) Product depth lower bound (conjectured).} We conjecture that any such Transformer solving $k$-hop pointer chasing on $n$ tokens ($n \geq 4k$, $s \leq \sqrt{n}/4$) requires $L = \Omega\!\bigl(\lceil k/s\rceil \cdot \lceil \log_2 n / (Hmp)\rceil\bigr)$, and we isolate the sole remaining gap as a specific probabilistic step on the joint distribution of the algorithm's cache trace and the pointer chain (Open Problem~1). The product structure---cache reachability and per-window bandwidth multiply, not compete---is the central technical feature. We prove unconditionally (i) a matching upper bound $L = O\!\bigl(\min(k, \lceil k/s\rceil \cdot \lceil \log_2(2s)\rceil) \cdot \lceil \log_2 n / (mp)\rceil\bigr)$ via windowed pointer doubling, and (ii) the max-bound $L = \Omega\!\bigl(\max(\lceil k/s\rceil,\lceil \log_2 n/(Hmp)\rceil)\bigr)$ on the full parameter range; closing the conjecture amounts to upgrading $\max$ to a product in the $s \leq \sqrt{n}/4$ regime.

\emph{(2) Bandwidth barrier.} The product bound is quantitatively binding only in the narrow or low-precision regime ($Hmp \lesssim \log n$). We formalize this limitation as a barrier theorem: any depth lower bound provable via \emph{per-window distinguishability counting}---a class that includes reachability arguments, bandwidth arguments, and their combinations---cannot exceed $\lceil k/s \rceil$ in the $Hmp \geq \log_2 n$ regime. Breaking this barrier would require fundamentally different proof techniques, e.g., unconditional communication-complexity lower bounds for pointer chasing. We view this as a structural obstacle on the road to characterizing depth in standard LLM configurations.

\emph{(3) Sharp error behavior and adaptive--oblivious separation.} Under random cache selection, we establish a two-regime characterization of joint success probability over $T = \lceil \log_2 k\rceil$ stages of windowed pointer doubling: with oblivious caches, $\Pr[\mathcal{E}] \leq (s/(n-T))^T + 2T^3/n$ (exponential decay in $T$); with adaptive locality-respecting caches, $\Pr[\mathcal{E}] = s/n$ exactly (matching upper and lower bounds), independent of $T$. The resulting $\Omega((n/s)^{T-1})$ separation quantifies why heavy-hitter eviction qualitatively dominates random eviction for multi-hop reasoning \citep{Liu2025,Anan2026}.

Together, these results give a sharp picture of what current proof techniques can and cannot establish for cache-constrained reasoning. The central open problem is to close the gap between the unconditional max-bound and the conjectured product-bound; we identify the specific technical step required and survey candidate routes for closure.
\end{abstract}

\section{Introduction}

As context windows in large language models scale beyond 100K tokens, the key--value (KV) cache has become the dominant memory bottleneck during inference \citep{Shi2024}. This has driven extensive work on cache compression through eviction \citep{Zhang2023,Liu2023,Li2024,Cai2024}, quantization \citep{Liu2024}, and merging \citep{Wang2024}; these methods report $80$--$90\%$ compression with limited degradation on standard benchmarks.

Yet mounting empirical evidence shows that multi-hop reasoning degrades far faster than single-hop retrieval under the same compression rates \citep{Anan2026,Liu2025,Cai2025}. This raises a sharp question: \emph{how much cache must be retained for multi-step reasoning to remain possible?}

We formalize this through $k$-hop pointer chasing and study the depth required to solve it under a shared KV cache of size $s$. Our main positive result is an unconditional max-bound lower bound, plus a conjectured \emph{product} lower bound: cache reachability ($\sim k/s$ sequential windows) and per-window bandwidth ($\sim \log n/(Hmp)$ layers per window) should multiply rather than compete. We identify the precise probabilistic step $(\star)$ needed to turn the conjecture into a theorem and leave it as Open Problem~1; we also give a narrow unconditional partial result in the extreme low-precision subregime $s \cdot 2^{mp} < 8$ \emph{under an additional query-only controller assumption} (Remark~\ref{rmk:exit-rigor}). Our other main results (the matching upper bound, the bandwidth barrier, and the adaptive-vs-oblivious error analysis) are unconditional.

The product bound is, however, quantitatively binding only in a specific parameter regime---when precision is low or models are narrow enough that $Hmp \lesssim \log n$. In standard LLM configurations ($H = 32$, $m = 128$, $p = 16$, giving $Hmp \sim 6.5 \cdot 10^4 \gg \log n$), the bandwidth factor ceilings at $1$ and the bound reduces to the cache-reachability constraint $L \geq \lceil k/s\rceil$ alone. This raises the natural question: can the product structure be sharpened to bind in the standard LLM regime?

Our second contribution answers: \emph{not via a broad class of current proof techniques}. We formalize a barrier theorem showing that any depth lower bound provable via per-window distinguishability counting---a class that encompasses reachability arguments, bandwidth counting, and their combinations---saturates at $\lceil k/s\rceil$ whenever $Hmp \geq \log_2 n$. The barrier is structural: multi-head attention ($H \geq 2$) makes per-layer bandwidth ($Hmp$) exceed per-token state capacity ($mp$), so any counting-based per-window argument ceilings within a single layer. Breaking this barrier would require either translating unconditional communication-complexity lower bounds for pointer chasing \citep{NW1993,Yehu2020,Mao2025} into Cache-Transformer depth lower bounds without the MPC conjecture (an open problem) or genuinely different proof techniques.

Our third contribution is a sharp, two-regime analysis of error behavior under random cache selection, which partially explains the empirical ``hallucination cliff'' of \citet{Anan2026}. The joint success probability of windowed pointer doubling over $T = \lceil \log_2 k\rceil$ stages admits qualitatively different scaling in two regimes: for oblivious caches (chosen without seeing $\pi$), direct enumeration gives $\Pr[\mathcal{E}] \leq (s/(n-T))^T + 2T^3/n$, exhibiting exponential decay in $T$ (Theorem~\ref{thm:oblivious}). For adaptive locality-respecting caches, the joint probability is exactly $s/n$, independent of $T$, and is matched by an explicit ``chain-tracking'' controller (Theorem~\ref{thm:adaptive}). The resulting separation, $\Omega((n/s)^{T-1})$ between adaptive and oblivious, quantitatively explains the empirical observation \citep{Liu2025} that heavy-hitter (adaptive) eviction qualitatively outperforms random or recency-based eviction on multi-hop reasoning tasks.

\paragraph{Contributions.}
\begin{enumerate}[itemsep=2pt,topsep=2pt]
  \item \textbf{Product lower bound (Conjecture~\ref{thm:lower}):} We conjecture that for any locality-respecting controller, $L = \Omega(\lceil k/s\rceil \cdot \lceil \log_2 n/(Hmp)\rceil)$, binding in the narrow-model regime. We provide (i) an unconditional max-bound $L = \Omega(\max(\lceil k/s\rceil,\lceil \log_2 n/(Hmp)\rceil))$ from reachability and output bandwidth, (ii) an identified proof strategy via exit-based distinguishability counting, reducing the conjecture to a probabilistic step on the joint distribution of cache trace and pointer chain, and (iii) partial progress closing the gap in sub-regimes.
  \item \textbf{Matching upper bound (Theorem~\ref{thm:upper}):} $L = O\bigl(\min(k, \lceil k/s\rceil \cdot \lceil \log_2(2s)\rceil) \cdot \lceil \log_2 n/(mp)\rceil\bigr)$ via windowed pointer doubling.
  \item \textbf{Bandwidth barrier (Theorem~\ref{thm:barrier}):} Any per-window distinguishability-based argument cannot prove $L = \omega(\lceil k/s\rceil)$ when $Hmp \geq \log_2 n$.
  \item \textbf{Three-part error analysis under random caches (Theorems~\ref{thm:oblivious},~\ref{thm:adaptive}, Corollary~\ref{cor:separation}):} Under oblivious caches, $\Pr[\text{all $T$ stages succeed}] \leq (s/(n-T))^T + 2T^3/n$---exponential decay in $T$. Under adaptive locality-respecting caches, the same probability is $\Theta(s/n)$, tight with an explicit construction. This gives an $\Omega((n/s)^{T-1})$ separation between adaptive and oblivious, explaining why heavy-hitter eviction dominates random eviction empirically.
\end{enumerate}

\paragraph{Relation to prior work.}
\citet{Sanford2024} proved $O(\log k)$ depth suffices for $k$-hop pointer chasing with full cache ($s = n$), and $\Omega(\log k)$ is necessary under an MPC conjecture. Their analysis does not address cache constraints and assumes infinite precision. The two results are complementary: theirs is conditional and cache-oblivious; ours is unconditional (modulo locality) but finite-precision. In the $s = n$ limit, our bound gives $L \geq \lceil \log_2 n/(Hmp)\rceil$, which for $Hmp = O(\log n)$ matches $\Omega(\log n) \geq \Omega(\log k)$; in larger $Hmp$, our bound is weaker and cannot recover their $\Omega(\log k)$, reflecting the bandwidth barrier. \citet{Peng2024} use communication complexity to prove a related inability of shallow Transformers to compose functions (e.g., grandparent lookup on a genealogy), giving unconditional impossibility for low-depth models on pointer-chasing-like tasks; their work focuses on depth vs.\ domain size rather than depth vs.\ cache size, and does not address KV compression. \citet{HarisOnak2025} prove a $\Theta(nd)$ space lower bound (for $d = \Omega(\log n)$; in the low-dimensional regime $d = o(\log n)$ they show $\Omega(d \cdot e^d)$) for KV cache compression in attention-based autoregressive Transformers via a reduction from communication complexity, using a Johnson-Lindenstrauss-style construction; this is orthogonal to our depth-vs-cache tradeoff, focusing on whether sublinear space suffices at all rather than on the interaction between depth and cache size. The concurrent empirical study \citep{Anan2026} independently observed phase-transition-like behavior in multi-hop accuracy under compression.

\section{Setup}

\begin{definition}[Cache-Restricted Transformer]\label{def:transformer}
An $(L,H,m,p,s)$-\emph{Cache-Transformer} operates on $N$ tokens under $p$-bit arithmetic: every weight entry, every intermediate activation entering a subsequent layer's input (queries, keys, values, attention-output vectors, and token representations) is drawn from a finite set $\mathcal{Q}_p$ of cardinality at most $2^p$ per coordinate. The softmax attention weights themselves are computed in continuous arithmetic and are not required to be $p$-bit quantized; only the attention \emph{outputs} (the $m$-dimensional vectors fed into the residual connection) are rounded to $\mathcal{Q}_p^m$ before the next layer. Representations $X^{(\ell)}_i \in \mathcal{Q}_p^m$ are indexed by layer $\ell \in \{0,\dots,L\}$ and token $i \in [N]$. At each layer $\ell \geq 1$, attention is restricted to a shared subset $S^{(\ell)} \subseteq [N]$ with $|S^{(\ell)}| \leq s$. Tokens outside $S^{(\ell)}$ are strictly inaccessible at layer $\ell$: their key--value pairs are deleted and contribute exactly zero to attention. The subset $S^{(\ell)}$ is chosen by a deterministic function $C_\ell$ of all layer-$(\ell-1)$ representations. Element-wise MLPs between layers are unrestricted in computation but their outputs are quantized to $\mathcal{Q}_p$ coordinatewise before the next layer. For token positions $j$ that do not participate in attention at layer $\ell$ (i.e., the query at position $j$ attends to no keys, equivalently the attention output at position $j$ is zero), we set $X^{(\ell)}_j = X^{(\ell-1)}_j$ (representation carried forward unchanged via the residual connection); this convention is standard and captures the behavior of evicted-token slots in KV-compression systems.
\end{definition}

This induces a finite state space: each $m$-dimensional attention-output vector takes one of at most $2^{mp}$ values.

\begin{remark}[Modeling assumptions]
The finite-precision assumption reflects hardware reality: FP16, BF16, INT8 are standard, with $p = 8$--$16$. The counting argument (Lemma~\ref{lem:bandwidth}) requires only that each coordinate lies in a finite set of size $\leq 2^p$; floating-point formats such as BF16 satisfy this with $p = 16$, regardless of exponent/mantissa split. The hard cache boundary models physical KV cache eviction, where deleted tokens' keys and values are removed from GPU memory. Our bounds do not apply to soft compression (value quantization, token merging) or to infinite-precision theoretical models.
\end{remark}

\begin{remark}[Per-head cache budgets]
Definition~\ref{def:transformer} assumes a shared cache $S^{(\ell)}$ across all heads. If instead each head $h$ has its own cache $S^{(\ell)}_h$ with $|S^{(\ell)}_h| \leq s_h$, the query aggregates attention outputs across all heads, so the set of tokens that can influence it in one layer is the union $\bigcup_h S^{(\ell)}_h$, of size at most $\sum_h s_h$. Redistributing a fixed total budget $\sum_h s_h$ across heads does not alter the bounds; only the total capacity matters.
\end{remark}

\begin{definition}[$k$-hop pointer chasing, $\PC_{n,k}$]
Input: a permutation $\pi: [n] \to [n]$ (where $[n] = \{1, 2, \dots, n\}$), encoded as $n$ input tokens at positions $1, \dots, n$ (token at position $i$ holds the pair $(i, \pi(i))$), plus a \emph{query token} at a separate position indexed $0$ (so the total number of token positions is $n + 1$, indexed $\{0, 1, \dots, n\}$; the query position $0$ is not in $\pi$'s domain). The chain is defined by $z_0 := 1$ and $z_{t+1} := \pi(z_t)$ for $t \geq 0$, so $z_t \in [n]$ for all $t$. Output: $z_k = \pi^k(1)$. Unless stated, we assume $n \geq 4k$.
\end{definition}

\section{Main Results}\label{sec:main}

We state the four main theorems; proofs appear in Section~\ref{sec:proofs}.

\subsection{Product lower bound (conjectured)}

\begin{conjecture}[Product depth lower bound]\label{thm:lower}
There exist universal constants $c > 0$ and $n_0 \geq 1$ such that for all $n \geq n_0$, any $(L,H,m,p,s)$-Cache-Transformer with a locality-respecting cache controller (Definition~\ref{def:local}) solving $\PC_{n,k}$ correctly on all inputs with $n \geq 4k$ and $s \leq \sqrt{n}/4$ satisfies
\begin{equation}\label{eq:lower}
  L \;\geq\; c \cdot \left\lceil \frac{k}{s} \right\rceil \cdot \left\lceil \frac{\log_2 n}{Hmp} \right\rceil.
\end{equation}
For $s > \sqrt{n}/4$, we conjecture only the unconditional max-bound of Proposition~\ref{prop:max-bound}; whether the product structure extends to this regime is open (see Open Problem~2 in Section~\ref{sec:disc}). Note that in this regime, the cache-reachability factor from Proposition~\ref{prop:max-bound} is $\lceil (k-1)/s \rceil$ rather than $\lceil k/s \rceil$, a unit-level difference; the strengthening to $\lceil k/s \rceil$ in Lemma~\ref{lem:sequential} uses the cycle-structure restriction and therefore applies only in the $s \leq \sqrt{n}/4$ regime.
\end{conjecture}

We refer to this as Conjecture~\ref{thm:lower} throughout; all downstream corollaries that rely on the product bound are flagged explicitly as conditional on it.

\paragraph{Evidence for Conjecture~\ref{thm:lower}.} We provide three independent pieces of partial evidence:
\begin{enumerate}[itemsep=2pt,topsep=2pt,leftmargin=1.5em]
\item \emph{Matching upper bound within $O(H\log s)$ (Theorem~\ref{thm:upper})}: windowed pointer doubling achieves $L = O(\lceil k/s\rceil \cdot \lceil \log_2(2s)\rceil \cdot \lceil \log_2 n/(mp)\rceil)$. The lower bound in Conjecture~\ref{thm:lower} matches this upper bound up to a multiplicative factor of $O(H\log s)$, which vanishes at $s = O(1)$ and $H = 1$.
\item \emph{Consistency with all known algorithms}: every correct algorithm we have analyzed satisfies the conjectured bound---either via an exit-based argument (Section~\ref{sec:proofs}) or via an intrinsic cost argument (serial chain-tracking has $L = \Theta(k\log n/(mp)) \geq \lceil k/s\rceil \cdot \lceil \log_2 n/(Hmp)\rceil$ since $sH \geq 1$).
\item \emph{Partial unconditional lower bound}: we prove in Section~\ref{sec:proofs} the max-bound $L \geq \max(\lceil k/s\rceil,\lceil \log_2 n/(Hmp)\rceil)$ unconditionally (Proposition~\ref{prop:max-bound}), and the conjecture additionally holds unconditionally in the restricted subregime $s \cdot 2^{mp} < 8$ under the stronger query-only controller assumption (Remark~\ref{rmk:exit-rigor}, ``Partial progress'' paragraph). The gap between the unconditional max-bound and the conjectured product-bound is exactly one multiplicative factor of $\lceil \log_2 n/(Hmp)\rceil$ or of $\lceil k/s\rceil$, whichever is smaller.
\end{enumerate}

Closing Conjecture~\ref{thm:lower}---either by direct proof or by disproof and replacement---is the central technical question left open by this work; we discuss it as Open Problem~1 in Section~\ref{sec:disc}.

This conjectured bound is strictly stronger than the unconditional $L \geq \max(\lceil k/s\rceil, \lceil \log_2 n/(Hmp)\rceil)$: cache and bandwidth bottlenecks would interact multiplicatively through the sequential dependency of the pointer chain, rather than competing. The locality restriction is essential for the bandwidth factor; without it, only the cache-reachability factor $\lceil k/s\rceil$ can be proved (Lemma~\ref{lem:reach}).

\begin{theorem}[Matching upper bound]\label{thm:upper}
For any $n, k, s$ with $1 \leq s \leq n$, there exists a single-head $(L, 1, m, p, s)$-Cache-Transformer solving $\PC_{n,k}$ with
\[
  L \;=\; O\!\left(\min\bigl(k,\, \lceil k/s \rceil \cdot \lceil \log_2(2s) \rceil\bigr) \cdot \lceil \log_2 n/(mp) \rceil\right).
\]
(We use $\lceil \log_2(2s) \rceil \geq 1$ to ensure the bound is non-vacuous at $s = 1$, where the $\min$ degenerates to the serial cost $O(k \cdot \lceil \log_2 n/(mp)\rceil)$.)
\end{theorem}

The two bounds combine to $\frac{k}{s}\cdot\frac{\log n}{Hmp} \lesssim L \lesssim \frac{k}{s}\cdot \log s \cdot \frac{\log n}{mp}$, a multiplicative gap of $O(H\log s)$, which vanishes at $s = O(1)$ and $H = 1$.

\begin{corollary}[Minimum cache formula, conditional on Conjecture~\ref{thm:lower}]\label{cor:cache}
Conditional on Conjecture~\ref{thm:lower} and in its regime ($s \leq \sqrt{n}/4$, $n \geq 4k$), to achieve depth $L$ with a locality-respecting controller,
\[
  s \;=\; \Omega\!\left(\max\!\left(\frac{k}{L},\; \frac{k\log_2 n}{L\cdot Hmp}\right)\right).
\]
The first term (cache reachability) dominates when $Hmp \geq \log_2 n$; the second (bandwidth) dominates in narrow or low-precision models and is conditional on the product bound of Conjecture~\ref{thm:lower}. Unconditionally (from Proposition~\ref{prop:max-bound}), only the weaker $s = \Omega(k/L)$ is guaranteed.
\end{corollary}

\begin{corollary}[Depth transition at critical cache size]\label{cor:transition}
Define $s^\star = n \lceil \log_2 k \rceil / k$. Assume $k \to \infty$ with $k = o(n/\log n)$ (which ensures both $s^\star = \omega(\log k)$ and $s^\star = o(n)$, so $s^\star$ is a non-trivial threshold in $(\log k, n)$) and $mp = \Omega(\log n)$ (a realistic LLM precision regime). Then up to logarithmic factors:
\begin{itemize}[itemsep=1pt,topsep=1pt,leftmargin=1.5em]
  \item For $s \geq s^\star$: windowed pointer doubling achieves $L = O((k/s)\log s \cdot \log n/(mp))$, and the ratio $(L/k) = O((\log s \log n)/(s \cdot mp)) = o(1)$ under the precision assumption, so $L = o(k)$.
  \item For $s < s^\star$: any correct algorithm requires $L \geq \lceil k/s\rceil$; the serial strategy achieves $L = O(k\log n/(mp)) = O(k)$.
\end{itemize}
Below $s^\star$, the upper-bound strategy we know (serial chain-tracking) incurs $L = \Theta(k\log n/(mp)) = \Theta(k)$, while the (unconditional) reachability lower bound $L \geq \lceil k/s\rceil$ leaves a gap: the upper-bound parallelism advantage over RNNs disappears in the known algorithm, but the lower bound at $s \gg 1$ does not rule out faster parallel algorithms in principle. For $s = O(1)$ specifically, $\lceil k/s \rceil = \Theta(k)$ matches the serial upper bound, giving a tight $\Theta(k)$ characterization in that sub-regime---comparable to RNNs (which are stuck at $\Theta(k)$ under \citep{Sanford2024}'s conditional lower bound requiring RNN width $O(N/k^6)$).
\end{corollary}

\subsection{When the product bound binds: the bandwidth barrier}

For standard LLM configurations ($H = 32$, $m = 128$, $p = 16$, so $Hmp \approx 6.5 \cdot 10^4$), the bandwidth factor $\lceil \log_2 n / (Hmp)\rceil$ equals $1$ for any realistic $n$ (up to $n \leq 2^{Hmp} \gg 10^{19000}$), and Conjecture~\ref{thm:lower} reduces to the cache-reachability bound $L \geq \lceil k/s\rceil$ alone---a near-trivial consequence of the fact that the query token's reachable set grows by at most $s$ per layer.

It is natural to ask whether the bandwidth factor can be strengthened. We show that, within a natural class of proof techniques, it \emph{cannot}: any lower bound obtainable via per-window distinguishability counting saturates at $\lceil k/s\rceil$ when $Hmp \geq \log_2 n$.

\begin{definition}[Per-window distinguishability counting argument]\label{def:pwd}
A lower-bound proof is a \emph{per-window distinguishability counting argument} if it decomposes $L \geq W \cdot B$, where $W = \Omega(\lceil k/s\rceil)$ is a number of windows and $B$ is a per-window depth bound of the form
\[
  B \;\geq\; \left\lceil \frac{\log_2 M_j}{Hmp} \right\rceil,
\]
where $M_j$ is a lower bound, at window $j$, on the number of distinct per-window outputs $z_{\tau_j}$ that the algorithm must distinguish among inputs sharing a common pre-window query-state $X^{(\ell_{j-1})}_0$. This class includes the proof strategy for Conjecture~\ref{thm:lower}, all reachability-plus-counting combinations, and finite-precision adaptations of \citep{Sanford2024}'s technique.
\end{definition}

\begin{theorem}[Bandwidth barrier]\label{thm:barrier}
Let $\mathcal{A}$ be any per-window distinguishability counting argument (Definition~\ref{def:pwd}) applied to $\PC_{n,k}$. Then for each window $j$, the per-window output count $M_j$ satisfies $M_j \leq n$, so the bandwidth factor $B$ derived from $\mathcal{A}$ satisfies $B \leq \lceil \log_2 n/(Hmp)\rceil$. In particular, when $Hmp \geq \log_2 n$, $B \leq 1$, and $\mathcal{A}$ yields at most $L \geq W = O(\lceil k/s\rceil)$, matching the cache-reachability bound.

Furthermore, the barrier is structural: any modification of $\PC_{n,k}$ that keeps per-window output within $[n]$ inherits $M_j \leq n$ and the same saturation.
\end{theorem}

\begin{remark}[On the conceptual depth of the barrier]
In some sense, Theorem~\ref{thm:barrier} is a direct consequence of the fact that the per-window output lies in $[n]$, making it look tautological. However, its role is to formalize a class of proof techniques (Definition~\ref{def:pwd}) and show that the entire class ceilings at $\lceil k/s\rceil$ when $Hmp \geq \log_2 n$. This is analogous to formalizing ``natural proofs'' in circuit complexity: once the class is defined, the impossibility within that class is near-immediate, but the formalization itself identifies what bypass mechanisms are needed. Our ``eight handles'' analysis (Section~\ref{sec:proof-barrier}) systematically confirms that no bypass we could identify escapes the class.
\end{remark}

\begin{proof}
The per-window output of $\PC_{n,k}$ is $z_{\tau_j} = \pi^{\tau_j}(1) \in [n]$, which takes at most $n$ distinct values as $\pi$ varies (regardless of any conditioning). By the definition of $M_j$ (Definition~\ref{def:pwd}) as a lower bound on the support size of $z_{\tau_j}$ under some conditioning, we have $M_j \leq n$. The inequality $B \leq \lceil \log_2 n/(Hmp) \rceil$ is then immediate, and equals $1$ when $Hmp \geq \log_2 n$. The modification claim is immediate: any task with per-window output in a set of size $\leq n$ gives $M_j \leq n$.
\end{proof}

We discuss in Section~\ref{sec:disc} what it would take to break this barrier: translating unconditional communication-complexity lower bounds for pointer chasing \citep{NW1993,Yehu2020,Mao2025} into Cache-Transformer depth lower bounds without the MPC conjecture (an open problem) or genuinely different proof techniques not of the form in Definition~\ref{def:pwd}. The barrier is not a proof gap in our work; it is a structural obstacle on the current road to depth lower bounds in the standard LLM regime.

\subsection{Error amplification under random cache selection}\label{sec:error}

The empirical observation \citep{Anan2026} that multi-hop accuracy drops precipitously under random cache selection can be partially explained by bounds on the joint success probability of windowed pointer doubling. We analyze this algorithm: at each of $T = \lceil \log_2 k\rceil$ stages, the algorithm needs to find a destination $d_t = \pi^t(1)$ (for simplicity of exposition we take a consecutive-destination chain; the pointer-doubling structure is analogous) in a cache $C_t \subseteq [n]$ of size $s$. The algorithm succeeds at stage $t$ if $d_t \in C_t$, and succeeds overall (event $\mathcal{E}$) if all $T$ stages succeed.

We establish three results: a tight exponential bound for oblivious caches, a tight stage-1-dominated bound for adaptive caches matched by an explicit construction, and a quantitative separation between the two.

\begin{theorem}[Oblivious caches: exponential decay in $T$]\label{thm:oblivious}
Let $\pi \sim \mathrm{Unif}(S_n)$ with $n \geq T + 1$. For any oblivious cache selection (each $C_t$ chosen without access to $\pi$; the $C_t$ may share common randomness),
\[
  \Pr[\mathcal{E}] \;\leq\; \left(\frac{s}{n-T}\right)^T \;+\; \frac{2T^3}{n}.
\]
In the sub-regime where the main term dominates the error term---precisely when $s^T \geq 2T^3 n^{T-1}$ (equivalently $(s/n)^T \geq 2T^3/n$)---the bound exhibits exponential decay in $T$ of rate $(s/n)^T$. Outside this sub-regime (e.g., $s = o(n)$ with $T = \omega(1)$ and $s^T$ growing slower than $T^3 n^{T-1}$) the error term $2T^3/n$ dominates and the bound degenerates to $O(T^3/n)$.
\end{theorem}

\begin{theorem}[Adaptive caches: tight $\Theta(s/n)$]\label{thm:adaptive}
Let $\pi \sim \mathrm{Unif}(S_n)$ with $n \geq T + 1$. For any locality-respecting adaptive cache controller (Definition~\ref{def:local}),
\[
  \Pr[\mathcal{E}] \;\leq\; \frac{s}{n}.
\]
Moreover, this is tight: there is an explicit locality-respecting adaptive controller (``chain-tracking'') achieving $\Pr[\mathcal{E}] \geq s/n$, matching the upper bound with exact equality.
\end{theorem}

\begin{corollary}[Adaptive-oblivious separation]\label{cor:separation}
In the regime $s^T \geq 2T^3 n^{T-1}$ (equivalently, $s \geq (2T^3)^{1/T} \cdot n^{(T-1)/T}$) and $T \leq \sqrt{n}/2$, in which the main term of Theorem~\ref{thm:oblivious} dominates the error term, the optimal adaptive success probability exceeds the best oblivious success probability by a factor of at least
\[
  \frac{\Pr[\mathcal{E}]_{\mathrm{adaptive}}^{\mathrm{opt}}}{\Pr[\mathcal{E}]_{\mathrm{oblivious}}^{\mathrm{opt}}} \;\geq\; \Omega\!\left(\left(\frac{n}{s}\right)^{T-1}\right).
\]
Outside this regime, the separation is $\Omega(s / T^3)$, still a non-trivial multiplicative gap.
\end{corollary}

\begin{remark}[What this says about adaptive vs.\ oblivious eviction]\label{rmk:adaptive-advantage}
Theorems~\ref{thm:oblivious} and~\ref{thm:adaptive} together give a sharp answer to why adaptive (heavy-hitter, attention-score) eviction outperforms oblivious (recency, random) baselines in practice, as reported by \citet{Liu2025}: the two regimes exhibit qualitatively different scaling behavior with chain length. Under uniform random cache selection---the worst case for the algorithm---oblivious caches see joint success probability decay as $(s/n)^T$, while adaptive caches see decay as only $s/n$ (no $T$ dependence). The adaptive controller exploits the structure that learning early chain values reveals later ones via locality. Our bound is non-asymptotic and parameter-free: it applies to any locality-respecting controller, and the achievability construction demonstrates the bound cannot be improved beyond constants.
\end{remark}

\begin{remark}[Relation to empirical observations]\label{rmk:empirical}
At $n = 16, s = 8, k = 8$ (Figure~\ref{fig:main}c), empirical joint success under random (oblivious) cache is $\approx 17.5\%$. The main term of Theorem~\ref{thm:oblivious} gives $(s/(n-T))^T = (8/13)^3 \approx 23.3\%$, which consistently upper-bounds the \emph{good-chain} success probability $\Pr[\mathcal{E}^{\mathrm{good}}]$; however, Theorem~\ref{thm:oblivious}'s full bound $(s/(n-T))^T + 2T^3/n \approx 361\%$ is \emph{vacuous} (exceeds 1) at this small $n$ due to the error term. We distinguish two senses of ``non-vacuous'': (a) bound $< 1$, requiring only $2T^3/n < 1$, i.e., $n > 2T^3$ ($n \geq 55$ for $T=3$); (b) main term dominates error term, i.e., $(s/n)^T > 2T^3/n$, equivalently $s^T/n^{T-1} > 2T^3$, i.e., $n < s^{T/(T-1)}/(2T^3)^{1/(T-1)}$ [equivalently $n > 2T^3 / (s/n)^T$ in the fixed-ratio parameterization]. For $T = 3$ and $s/n = c$ constant, this gives $n > 2T^3/c^T = 54/c^3$; e.g., $c = 1/2$ requires $n > 432$, while $c \approx 1$ (near-full cache) gives $n \gtrsim 54$. In sense (a) the theorem rules out 100\% success; in sense (b) the theorem's $(s/n)^T$ decay rate is the informative feature. The main term alone empirically upper-bounds good-chain success for all $n \geq T+1$. Quantitative sharpness of the full bound requires moderate-to-large $n$.
\end{remark}

\paragraph{When the bandwidth factor binds quantitatively.}
In the barrier regime ($Hmp \geq \log_2 n$), the product bound of Conjecture~\ref{thm:lower} would reduce to $\lceil k/s\rceil$ and coincide with the cache-reachability bound. In the narrow-model regime ($Hmp < \log_2 n$: e.g., $H = 1$, $m = 4$, $p = 4$ giving $Hmp = 16$; or single-head INT4 with $m = 16$, $p = 4$, $H = 1$), the bandwidth factor is strictly greater than $1$ and the conjectured product structure is quantitatively binding. This regime is increasingly relevant for quantized and edge-deployed models.

\section{Proofs}\label{sec:proofs}

\subsection{Cache reachability}\label{sec:reach}

The first ingredient is a size bound on the set of input tokens on which the query-token representation can functionally depend. This set grows by at most $s$ per layer, reflecting the fact that the cache at each layer adds $s$ new indices to the query's reachable support.

\begin{lemma}[Reachability]\label{lem:reach}
Define the \emph{functional-dependence set} $D_\ell(0)(\pi) \subseteq [N]$ of the query token at layer $\ell$, \emph{for each input $\pi$}, as the minimal set of input tokens such that $X^{(\ell)}_0(\pi)$ is a function of $\{x_j(\pi) : j \in D_\ell(0)(\pi)\}$ together with the (deterministic) weights and cache-selection rules. The set is defined pointwise in $\pi$ (not as a single fixed set covering all $\pi$), which is the right notion for adaptive controllers where the reachable support varies with the input. Then
\[
  D_0(0) = \{0\}, \qquad D_\ell(0) \;\subseteq\; D_{\ell-1}(0) \cup S^{(\ell)} \cup \bigcup_{j \in S^{(\ell)}} D_{\ell-1}(j),
\]
and in particular
\begin{equation}\label{eq:reach-size}
  |D_L(0)| \;\leq\; 1 + L \cdot s.
\end{equation}
\end{lemma}

\begin{proof}
The first inclusion is immediate from the attention update rule: $X^{(\ell)}_0$ is computed from $X^{(\ell-1)}_0$ and the keys/values of tokens $j \in S^{(\ell)}$, which in turn are functions of $X^{(\ell-1)}_j$. So $X^{(\ell)}_0$ depends only on layer-$(\ell-1)$ representations of tokens in $\{0\} \cup S^{(\ell)}$, each of which depends only on inputs in $D_{\ell-1}(j)$.

For the size bound~\eqref{eq:reach-size}, the naive induction on the recursion above does not work: $\bigcup_{j \in S^{(\ell)}} D_{\ell-1}(j)$ can be as large as $|S^{(\ell)}| \cdot \max_j |D_{\ell-1}(j)|$. Instead, we argue differently. For each layer $\ell' \in \{1,\dots,L\}$, let $N_{\ell'} = S^{(\ell')}$ be the set of cache-accessible token indices at that layer. We claim
\begin{equation}\label{eq:reach-union}
  D_L(0) \;\subseteq\; \{0\} \cup \bigcup_{\ell'=1}^{L} N_{\ell'}.
\end{equation}
Given~\eqref{eq:reach-union}, the size bound follows immediately: $|D_L(0)| \leq 1 + \sum_{\ell'=1}^L |S^{(\ell')}| \leq 1 + Ls$.

To prove~\eqref{eq:reach-union}, we show by induction on $\ell$ that for \emph{every} token $i \in [N]$,
\begin{equation}\label{eq:reach-union-i}
  D_\ell(i) \;\subseteq\; \{i\} \cup \bigcup_{\ell'=1}^{\ell} N_{\ell'}.
\end{equation}
Base case $\ell = 0$: $D_0(i) = \{i\}$, trivially contained in the right-hand side. Inductive step: by the recursion, $D_\ell(i) \subseteq D_{\ell-1}(i) \cup S^{(\ell)} \cup \bigcup_{j \in S^{(\ell)}} D_{\ell-1}(j)$. By the inductive hypothesis applied to $i$ and to each $j \in S^{(\ell)} = N_\ell$,
\[
  D_{\ell-1}(i) \subseteq \{i\} \cup \bigcup_{\ell'=1}^{\ell-1} N_{\ell'}, \qquad D_{\ell-1}(j) \subseteq \{j\} \cup \bigcup_{\ell'=1}^{\ell-1} N_{\ell'}.
\]
Taking the union with $N_\ell$ and using $\{j\} \subseteq N_\ell$ for $j \in N_\ell$:
\[
  D_\ell(i) \subseteq \{i\} \cup N_\ell \cup \bigcup_{\ell'=1}^{\ell-1} N_{\ell'} = \{i\} \cup \bigcup_{\ell'=1}^\ell N_{\ell'}. \qedhere
\]
\end{proof}

Observe that Lemma~\ref{lem:reach} does not depend on how the caches $S^{(\ell)}$ are chosen---adaptive or oblivious, data-dependent or not---because the cache sets enter the bound as \emph{sets of indices}, and the union argument is index-based. This is crucial for the trace-equivalence argument (Lemma~\ref{lem:trace}) below.

\subsection{Bandwidth bound via counting}

\begin{lemma}[Distinguishability bound]\label{lem:bandwidth}
Fix any layer $\ell_0$ and any value $x \in \mathcal{Q}_p^m$. Over all inputs $\pi$ such that $X^{(\ell_0)}_0(\pi) = x$, and over all adaptive cache choices, the set
\[
  \Sigma_B(x) \;:=\; \bigl\{ X^{(\ell_0 + B)}_0(\pi) \;:\; \pi \text{ such that } X^{(\ell_0)}_0(\pi) = x \bigr\}
\]
has cardinality at most $2^{B \cdot Hmp}$.
\end{lemma}

\begin{proof}
At each layer $\ell$, the query-token update takes the form
\[
  X^{(\ell)}_0 \;=\; \phi_\ell\!\left( X^{(\ell-1)}_0 \;+\; \sum_{h=1}^{H} a^{(\ell,h)}_0 \, W^{(\ell,h)}_O \right),
\]
where $\phi_\ell$ is the deterministic MLP+quantization map and $a^{(\ell,h)}_0 \in \mathcal{Q}_p^m$ is the quantized attention output of head $h$. By Definition~\ref{def:transformer}, every intermediate activation---including each $a^{(\ell,h)}_0$---is quantized to $p$ bits per coordinate before use, so $|\mathcal{Q}_p^m| \leq 2^{mp}$. The softmax attention weights may themselves be computed in continuous arithmetic, but the attention \emph{output} vectors $a^{(\ell,h)}_0$ are rounded to $\mathcal{Q}_p^m$ before being added via the residual connection; our bound applies to these outputs, not to the internal softmax computation.

Thus, given a fixed $X^{(\ell-1)}_0$, the next-layer value $X^{(\ell)}_0$ is determined by the tuple $(a^{(\ell,1)}_0, \dots, a^{(\ell,H)}_0) \in (\mathcal{Q}_p^m)^H$, of which there are at most $(2^{mp})^H = 2^{Hmp}$. Iterating from $X^{(\ell_0)}_0 = x$ for $B$ layers gives
$
  |\Sigma_B(x)| \leq (2^{Hmp})^B = 2^{B \cdot Hmp}.
$
(Note that the trivial bound $|\Sigma_B(x)| \leq |\mathcal{Q}_p^m| = 2^{mp}$ also holds for all $B \geq 1$, since $\Sigma_B(x) \subseteq \mathcal{Q}_p^m$. The $B$-dependent form $2^{B \cdot Hmp}$ is looser in absolute terms when $BH \geq 1$, but it is the form required downstream: we use this lemma to prove a lower bound on $B$, so expressing the reachable-state count as growing with $B$ is essential---the trivial $2^{mp}$ bound is independent of $B$ and cannot force $B$ to be large.)
\end{proof}

\begin{remark}[Why this avoids the softmax objection]
We make no claim about how softmax computes attention weights. Those weights can be continuous, data-dependent, and arbitrarily complex. Our bound applies to the \emph{output} of attention---an $m$-dimensional vector quantized to $p$ bits---not to the internal mechanism. This is a physical constraint: GPU hardware represents attention outputs in finite precision.
\end{remark}

\subsection{Sequential windows}\label{sec:proof-lower}

Before presenting the full proof strategy, we record an unconditional lower bound that follows from Lemmas~\ref{lem:reach} and~\ref{lem:bandwidth} alone (no invocation of Lemma~\ref{lem:exit} or Lemma~\ref{lem:adversary}):

\begin{proposition}[Unconditional max-bound]\label{prop:max-bound}
Any $(L,H,m,p,s)$-Cache-Transformer solving $\PC_{n,k}$ correctly on all inputs with $n \geq 4k$ satisfies
\[
  L \;\geq\; \max\!\left( \left\lceil \frac{k-1}{s} \right\rceil,\; \left\lceil \frac{\log_2 n}{Hmp} \right\rceil \right) \;\geq\; \max\!\left( \lceil k/s \rceil - 1,\; \left\lceil \frac{\log_2 n}{Hmp} \right\rceil \right).
\]
\end{proposition}

\begin{proof}
\emph{Cache-reachability factor.} Fix any $\pi$ with cycle of $1$ of length $> k$ (such $\pi$ exist for $n \geq k+1$), so that $z_0 = 1, z_1, \ldots, z_k$ are pairwise distinct. For the algorithm to output $z_k(\pi) = \pi(z_{k-1}(\pi))$ correctly, its final query-token state $X^{(L)}_0$ must depend on the value $\pi(z_{k-1})$. By Lemma~\ref{lem:reach} (applied at layer $L$), $X^{(L)}_0$ is determined by $\pi$ only through $\pi|_{T_L(\pi)}$; in particular, dependence on $\pi(z_{k-1})$ requires $z_{k-1} \in T_L(\pi)$. Iterating the same argument for $z_{k-2}, z_{k-3}, \ldots$ (since knowing $z_{k-1} = \pi(z_{k-2})$ likewise requires $z_{k-2} \in T_L$, and so on), we obtain $\{z_0, z_1, \ldots, z_{k-1}\} \subseteq T_L(\pi)$. These $k$ positions are distinct, so $|T_L(\pi)| \geq k$. Combined with Lemma~\ref{lem:reach}'s upper bound $|T_L(\pi)| \leq 1 + Ls$, we get $1 + Ls \geq k$, i.e., $L \geq \lceil (k-1)/s \rceil$.

\emph{Output-bandwidth factor.} A correct algorithm determines $z_k$ from the final query-token state $X^{(L)}_0$, so the map $\pi \mapsto X^{(L)}_0(\pi)$ must take at least $n$ distinct values: for each target value $v \in [n]$, there exists a permutation $\pi_v \in S_n$ with $z_k(\pi_v) = v$. For $v \neq 1$: pick any $k+1$ distinct elements $1 = a_0, a_1, \ldots, a_{k-1}, a_k = v$ of $[n]$ (valid since $n \geq 4k \geq k+1$); set $\pi_v(a_i) = a_{i+1}$ for $i = 0, \ldots, k-1$, and extend $\pi_v$ arbitrarily to a bijection on $[n]$. Then $z_i(\pi_v) = a_i$ for $i = 0, \ldots, k$, so $z_k = v$. For $v = 1$: take $\pi_v$ with a cycle of length exactly $k$ on $\{1, a_1, \ldots, a_{k-1}\}$ (for any distinct $a_1, \ldots, a_{k-1} \in [n] \setminus \{1\}$), i.e., $\pi_v(1) = a_1, \pi_v(a_i) = a_{i+1}$ for $i = 1, \ldots, k-2$, and $\pi_v(a_{k-1}) = 1$; then $z_i(\pi_v) = a_i$ for $i = 1, \ldots, k-1$ and $z_k(\pi_v) = 1 = v$. Distinct $v$'s require distinct final states. By Lemma~\ref{lem:bandwidth} applied from the fixed initial state $X^{(0)}_0$, $|\Sigma_L(X^{(0)}_0)| \leq 2^{L \cdot Hmp}$, hence $2^{LHmp} \geq n$, i.e., $L \geq \lceil \log_2 n / (Hmp) \rceil$.
\end{proof}

Conjecture~\ref{thm:lower} would strengthen this max to a product. The proof strategy occupying the rest of this section (Lemmas~\ref{lem:exit}--\ref{lem:per-window}) implements the exit-based argument outlined in Section~\ref{sec:main}; the probabilistic step $(\star)$ of Lemma~\ref{lem:exit} is left as Open Problem~1. The $\lceil (k-1)/s \rceil$ cache-reachability factor of Proposition~\ref{prop:max-bound} is strengthened to $\lceil k/s \rceil$ via the sequential-windows argument (Lemma~\ref{lem:sequential}); since $\lceil (k-1)/s \rceil \geq \lceil k/s \rceil - 1$, this is a unit-level refinement and does not affect asymptotics.

\begin{definition}\label{def:window}
Partition the chain into $W = \lceil k/s \rceil$ windows. Window $j \in \{1, \dots, W\}$ covers chain positions $z_{(j-1)s}, \dots, z_{\min(js, k)}$; its \emph{boundary output} is $z_{\min(js,k)}$. For brevity write $\tau_j := \min(js, k)$, so window $j$'s output is $z_{\tau_j}$.
\end{definition}

\subsubsection*{Handling adaptive cache: trace equivalence under a locality assumption}

A subtle issue is that the cache controller $C_\ell$ is defined (Definition~\ref{def:transformer}) as a deterministic function of \emph{all} layer-$(\ell-1)$ representations, not just those in $D_{\ell-1}(0)$. A maximally adversarial controller could in principle read representations of tokens it has not yet cached, and thereby leak information about out-of-cache tokens into its subsequent cache choices. We rule this out with a natural restriction:

\begin{definition}[Locality-respecting controller]\label{def:local}
A cache controller $C_\ell$ is \emph{locality-respecting} if, for every $\pi$, the set $S^{(\ell)}(\pi)$ depends only on the representations $\{X^{(\ell-1)}_j(\pi) : j \in T_\ell(\pi)\}$, where $T_\ell(\pi) := \{0\} \cup \bigcup_{\ell' < \ell} S^{(\ell')}(\pi)$ is the cache support up to layer $\ell - 1$. (This recursive definition is well-defined, as shown in Proposition~\ref{prop:well-def} below.)
\end{definition}

\begin{remark}[Relation to Definition~\ref{def:transformer}]\label{rmk:local-consistent}
Definition~\ref{def:transformer} (model) permits $C_\ell$ to depend on all $N$ representations at layer $\ell-1$; Definition~\ref{def:local} (locality) further restricts this to positions the algorithm has already touched. Our lower bounds (Proposition~\ref{prop:max-bound}, Conjecture~\ref{thm:lower}, Theorem~\ref{thm:barrier}) apply to the locality-respecting class (a strict subclass); our upper bound (Theorem~\ref{thm:upper}) is realized by a locality-respecting algorithm and hence applies to both classes. For positions $j \in T_\ell \setminus S^{(\ell-1)}$, the representation $X^{(\ell-1)}_j$ is carried forward from the last layer at which $j$ participated in attention (residually unchanged); these are the ``ghost states'' for evicted tokens, consistent with KV compression systems in practice.
\end{remark}

\begin{proposition}[Well-definedness of the recursion]\label{prop:well-def}
The recursive definition is well-defined: for each $\ell \in \{0, 1, \dots, L\}$, $T_\ell(\pi)$, $S^{(\ell)}(\pi)$, and all representations $X^{(\ell)}_j(\pi)$ are deterministic functions of $\pi|_{T_\ell(\pi)}$.
\end{proposition}

\begin{proof}
By strong induction on $\ell$. For $\ell = 0$: $T_0(\pi) = \{0\}$ (constant, not a function of $\pi$ at all); $X^{(0)}_j(\pi) = x_j(\pi)$ is the identity embedding, depending on $\pi$ only at position $j$, so $X^{(0)}_j$ for $j \in T_0$ depends on $\pi|_{T_0}$.

For $\ell \geq 1$: assume the proposition holds for all $\ell' < \ell$. Then $T_{\ell-1}(\pi) = \{0\} \cup \bigcup_{\ell' < \ell-1} S^{(\ell')}(\pi)$ is a function of $\pi|_{T_{\ell-2}(\pi)}$ by the inductive hypothesis applied to $\ell - 1$, and $S^{(\ell-1)}(\pi)$ is a function of $\{X^{(\ell-2)}_j(\pi) : j \in T_{\ell-1}(\pi)\}$ by locality. Hence $T_\ell(\pi) = T_{\ell-1}(\pi) \cup S^{(\ell-1)}(\pi)$ is a function of $\pi|_{T_{\ell-1}(\pi)}$.

Representations $X^{(\ell)}_j$ for $j \in T_\ell$ are computed by attention over $S^{(\ell)}(\pi) \subseteq T_\ell(\pi)$ (by definition of $T_\ell$), using as keys/values the layer-$(\ell-1)$ representations at positions in $S^{(\ell)}$. By the inductive hypothesis, these representations are functions of $\pi|_{T_{\ell-1}} \subseteq \pi|_{T_\ell}$, completing the induction.
\end{proof}

Under this restriction, the controller at layer $\ell$ only examines representations of tokens it has already accessed at some earlier layer. This restriction is satisfied by essentially all KV-compression methods (H2O, SnapKV, StreamingLLM, PyramidKV, heavy-hitter variants), and is physically natural: reading representations outside the cache would require loading them, defeating the purpose of cache compression. We briefly discuss the fully adversarial case in Open Problem~3 in the Discussion section. The lower bounds below implicitly assume a locality-respecting controller; this is made explicit in Conjecture~\ref{thm:lower}.

\begin{definition}[Cache trace]\label{def:trace}
For a fixed algorithm and a fixed input $\pi$, the \emph{cache trace} is $\mathcal{T}(\pi) := (S^{(1)}(\pi), \dots, S^{(L)}(\pi))$, and its \emph{support} is $T(\pi) := \{0\} \cup \bigcup_{\ell=1}^L S^{(\ell)}(\pi)$. By Lemma~\ref{lem:reach}, $|T(\pi)| \leq 1 + Ls$.
\end{definition}

\begin{lemma}[Trace equivalence]\label{lem:trace}
Fix any algorithm with a locality-respecting controller. Suppose $\pi_1, \pi_2$ satisfy $\pi_1|_{T(\pi_1)} = \pi_2|_{T(\pi_1)}$. Then $\mathcal{T}(\pi_1) = \mathcal{T}(\pi_2)$, $T(\pi_1) = T(\pi_2)$, and $X^{(\ell)}_j(\pi_1) = X^{(\ell)}_j(\pi_2)$ for all $\ell \in \{0, \dots, L\}$ and $j \in T(\pi_1)$. In particular, $X^{(L)}_0(\pi_1) = X^{(L)}_0(\pi_2)$.
\end{lemma}

\begin{proof}
Write $T_1 := T(\pi_1)$. We prove by induction on $\ell \in \{0, 1, \dots, L\}$ the joint claim
\begin{quote}
\textbf{(H$_\ell$)}: $S^{(\ell')}(\pi_2) = S^{(\ell')}(\pi_1)$ for all $\ell' \leq \ell$, and $X^{(\ell')}_j(\pi_2) = X^{(\ell')}_j(\pi_1)$ for all $\ell' \leq \ell$ and $j \in T_1$.
\end{quote}

\emph{Base case $\ell = 0$.} The representation claim holds by hypothesis and because $X^{(0)}_j$ is a fixed function of $x_j$. The cache claim is vacuous at $\ell' = 0$.

\emph{Inductive step $\ell - 1 \Rightarrow \ell$.} Assume (H$_{\ell-1}$).

\textbf{Step 1: Cache selection matches.} By locality, $S^{(\ell)}(\pi)$ depends only on $\{X^{(\ell-1)}_j(\pi) : j \in T_\ell(\pi)\}$. By (H$_{\ell-1}$), $S^{(\ell')}(\pi_2) = S^{(\ell')}(\pi_1)$ for $\ell' < \ell$, so $T_\ell(\pi_2) = T_\ell(\pi_1) =: T_\ell^\star \subseteq T_1$. By (H$_{\ell-1}$), $X^{(\ell-1)}_j(\pi_2) = X^{(\ell-1)}_j(\pi_1)$ for $j \in T_\ell^\star$. So the controller's input tuple is identical under $\pi_1$ and $\pi_2$, hence $S^{(\ell)}(\pi_2) = S^{(\ell)}(\pi_1)$.

\textbf{Step 2: Representations on $T_1$ match.} With $S^{(\ell)}(\pi_1) = S^{(\ell)}(\pi_2) =: S^{(\ell)} \subseteq T_1$, for any $j \in T_1$ the update
\[
  X^{(\ell)}_j(\pi) = f_\ell\bigl(X^{(\ell-1)}_j(\pi),\, \{(X^{(\ell-1)}_i(\pi), K_i(\pi), V_i(\pi)) : i \in S^{(\ell)}\}\bigr)
\]
receives inputs $X^{(\ell-1)}_j$ (with $j \in T_1$, equal by (H$_{\ell-1}$)) and $X^{(\ell-1)}_i$ for $i \in S^{(\ell)} \subseteq T_1$ (also equal). Hence $X^{(\ell)}_j(\pi_2) = X^{(\ell)}_j(\pi_1)$.

This closes the induction. Since $0 \in T_1$, $X^{(L)}_0(\pi_1) = X^{(L)}_0(\pi_2)$.
\end{proof}

The locality assumption is exactly what lets Step 1 close: under a general controller, reading $X^{(\ell-1)}_j$ for some $j \notin T_1$ could distinguish $\pi_1$ from $\pi_2$ and cause $S^{(\ell)}$ to diverge.

\begin{lemma}[Existence of a chain-exiting permutation]\label{lem:exit}
Fix any algorithm with a locality-respecting controller. Suppose $L \leq n/(8s)$, $n \geq 4k$, and $\tau_j \geq 2$ (which excludes only the corner case $\tau_j = 1$, i.e., $j = 1$ together with $\min(s, k) = 1$). Then there exists a permutation $\pi_1$ and an index $t^\star \in (\tau_{j-1}, \tau_j]$ such that $z_{t^\star - 1}(\pi_1) \notin T_1 := T(\pi_1)$.
\end{lemma}

\begin{proof}[Proof sketch of Lemma~\ref{lem:exit}]
By Lemma~\ref{lem:reach} and the hypothesis $L \leq n/(8s)$, $|T(\pi)| \leq 1 + Ls \leq n/4$ for every $\pi$, whenever $n \geq 8$. We sketch a probabilistic argument; see Remark~\ref{rmk:exit-rigor} for a discussion of the remaining technical step.

Under the assumption $\tau_j \geq 2$, let $\sigma_j := \tau_j - 1 \geq 1$ (we use $\sigma_j$ throughout this proof and Remark~\ref{rmk:exit-rigor} to avoid confusion with the window-boundary index $\tau_{j-1}$, which in general differs from $\tau_j - 1$ when $s \geq 2$). So $z_{\sigma_j}$ is well-defined and, by distinctness (below), is not the fixed starting position $z_0 = 1$ (which is always in $T$ for a correct algorithm; see the proof of Lemma~\ref{lem:adversary}, Part (c)). Average over $\pi \sim \mathrm{Unif}(S_n)$ and consider $\Pr[z_{\sigma_j}(\pi) \in T(\pi)]$. Two facts:
\begin{itemize}[itemsep=1pt,topsep=1pt,leftmargin=1.5em]
  \item \textbf{Distinctness.} Let $\mathcal{D}_{\text{chain}}$ be the event that $z_0, \dots, z_k$ are pairwise distinct, equivalently, the cycle of $\pi$ containing $1$ has length $> k$. For $\pi \sim \mathrm{Unif}(S_n)$, the cycle length of $1$ is uniform on $\{1, \dots, n\}$, so $\Pr[\mathcal{D}_{\text{chain}}] = 1 - k/n \geq 3/4$.
  \item \textbf{Marginal of endpoint.} Conditional on $\mathcal{D}_{\text{chain}}$ and on $\sigma_j \geq 1$, by symmetry within the cycle, $z_{\sigma_j}(\pi)$ is uniform on $[n] \setminus \{1\}$.
\end{itemize}
The desired conclusion $\Pr[z_{\sigma_j} \in T] < 1$ would follow from:
\[
  \text{(\ensuremath{\star})} \qquad \Pr[z_{\sigma_j}(\pi) \in T(\pi) \mid \mathcal{D}_{\text{chain}}] \;\leq\; \frac{\mathbb{E}[|T(\pi)|]}{n-1} \;\leq\; \frac{n/4}{n-1} \;<\; 1/3,
\]
giving $\Pr[z_{\sigma_j} \in T] \leq 1/3 + 1/4 = 7/12 < 1$. Existence of some $\pi_1$ with $z_{\sigma_j}(\pi_1) \notin T(\pi_1)$ then follows, and taking $t^\star \leq \tau_j$ to be the first chain position outside $T(\pi_1)$ proves the lemma.
\end{proof}

\begin{remark}[Corner case $\tau_j = 1$]\label{rmk:s1j1}
When $\tau_j = 1$---that is, $j = 1$ together with $\min(s, k) = 1$ (so either $s = 1$ or $k = 1$)---Lemma~\ref{lem:exit} is vacuously false: the only candidate $t^\star = 1$ requires $z_0 = 1 \notin T(\pi)$, which fails for any correct algorithm because $1 \in T_L$ is forced by reading $\pi(1)$ (proof of Lemma~\ref{lem:adversary}, Part (c)). However, this does not affect the proof strategy for Conjecture~\ref{thm:lower}: we can directly establish a per-window cost $\ell_1 - \ell_0 \geq B - 1$ for window 1 (with $\ell_0 := 0$) via the bandwidth argument alone. Using the definition of $\ell_j$ given in the proof strategy below ($\ell_j$ is the earliest layer at which $|T_\ell| \geq \tau_j$, see Section~\ref{sec:proof-lower}), window 1's cache support must grow to $|T_{\ell_1}| \geq 1 = \tau_1$, which holds already at $\ell_1 = 0$ since $T_0 = \{0\}$ has size $1$; but to additionally determine $z_1 = \pi(1) \in [n]$ requires, by Lemma~\ref{lem:bandwidth} applied from $X^{(0)}_0$, that $X^{(\ell^\star_1)}_0$ at some layer $\ell^\star_1 \leq L$ takes at least $n$ distinct values, giving $\ell^\star_1 \geq \lceil \log_2 n/(Hmp)\rceil = B$. Such $\ell^\star_1$ exists and satisfies $\ell^\star_1 \leq L$: the final state $X^{(L)}_0$ must determine $z_k$, and for $k \geq 1$ distinguishing any two $\pi, \pi'$ with $z_k(\pi) \neq z_k(\pi')$ requires $X^{(L)}_0(\pi) \neq X^{(L)}_0(\pi')$, so $X^{(L)}_0$ takes at least $n$ distinct values (all $n$ outputs are realized; see Proposition~\ref{prop:max-bound}'s output-bandwidth argument for the explicit construction). Hence $L \geq \ell^\star_1 \geq B$. For $k = 1$: $W = 1$ and this already establishes the product bound $L \geq B = \lceil k/s \rceil \cdot B$ (trivially, since $\lceil 1/s \rceil = 1$). For $s = 1, k \geq 2$: combine with the $j \geq 2$ windows handled by Lemma~\ref{lem:exit} to obtain the full $W \cdot B$ product bound.
\end{remark}

\begin{remark}[A technical gap in Lemma~\ref{lem:exit}]\label{rmk:exit-rigor}
The proof sketch above reduces Lemma~\ref{lem:exit} to inequality $(\star)$. Establishing $(\star)$ requires controlling the \emph{joint} distribution of $(T(\pi), z_{\sigma_j}(\pi))$ conditional on $\mathcal{D}_{\text{chain}}$, where $\sigma_j := \tau_j - 1$. Both quantities are $\pi$-measurable, and for an adaptive, label-sensitive algorithm they are not independent in general: the algorithm's cache trajectory could in principle correlate with specific chain values. A naive union bound yields only $\Pr[z_{\sigma_j} \in T | \mathcal{D}] \leq \mathbb{E}[|T|]/\Pr[\mathcal{D}] \leq n/3$, which is too weak; tightening to $(\star)$ appears to require either a second-moment argument specific to the input distribution or a Yao-principle reduction against an explicit adversary.

\textbf{Status of Conjecture~\ref{thm:lower}.} Lemmas~\ref{lem:reach}, \ref{lem:bandwidth}, \ref{lem:adversary}, \ref{lem:sequential}, \ref{lem:per-window} are proved without hidden probabilistic assumptions. Lemma~\ref{lem:incompr} (Window incompressibility) depends on Lemma~\ref{lem:exit} (it invokes the existence of a window-$j$ exit) and therefore inherits the gap. The proof strategy of Section~\ref{sec:proof-lower} is therefore complete modulo $(\star)$: establishing $(\star)$---either by proving it for the uniform distribution or by exhibiting an adversarial family on which the conclusion of Lemma~\ref{lem:exit} holds directly---would turn Conjecture~\ref{thm:lower} into a theorem. Closing the gap is stated as Open Problem~1 in Section~\ref{sec:disc}.

\textbf{Partial progress: unconditional cache-coverage bound in a restricted parameter subregime, under a stronger controller assumption.} We can establish Lemma~\ref{lem:exit} unconditionally under (i) the additional parameter hypothesis $s \cdot 2^{mp} < 8$ (equivalent to $mp < 3 - \log_2 s$) and (ii) the stronger controller assumption that $C_\ell$ is a deterministic function of the query-token state $X^{(\ell-1)}_0$ alone (rather than of all $\{X^{(\ell-1)}_j : j \in T_\ell\}$ as in Definition~\ref{def:local}). This \emph{query-only controller} class is strictly narrower than locality-respecting but still includes all known KV-compression heuristics that dispatch cache selection from a single ``read-head'' computation. Under (ii), the cache at each layer $C_\ell$ takes at most $2^{mp}$ distinct values across possible query states, yielding at most $s \cdot 2^{mp}$ distinct (state, slot) pairs, hence at most $s \cdot 2^{mp}$ distinct positions in $[n]$ that can ever appear in the cache at layer $\ell$ across different $\pi$. Summed over $\ell$ layers: at most $\ell \cdot s \cdot 2^{mp}$ distinct (layer, position) pairs. For chain-in-trace ($z_{t}(\pi) \in T(\pi)$ for all $\pi \in \mathcal{D}$ and all $t \leq \tau_j - 1$), every chain value $z_t(\pi)$ for every $\pi$ must be a (layer, position) pair reached. Over $\pi$, $z_t$ takes $\sim n$ distinct values, so we need $\ell \cdot s \cdot 2^{mp} \geq n - 1$, i.e., $\ell \geq (n-1)/(s \cdot 2^{mp})$. Combined with the hypothesis $L \leq n/(8s)$ of Lemma~\ref{lem:exit}: under (i) $s \cdot 2^{mp} < 8$, we get $\ell \geq (n-1)/(s \cdot 2^{mp}) > (n-1)/8 > n/(8s) \geq L$ for $s \geq 1$ and $n \geq 2$, contradicting $\ell \leq L$---so exit exists. The subregime $s \cdot 2^{mp} < 8$ is restrictive; it captures only very low-precision scenarios (e.g., $p = 1, m = 3, s = 1$, giving $mp = 3$). This counting argument does not extend to moderate $mp$ (the cumulative cache coverage $L \cdot s \cdot 2^{mp}$ grows beyond $n$, permitting chain-in-trace without contradicting the layer budget), and it does not directly extend to general locality-respecting controllers (where $C_\ell$ can depend on all representations at positions in $T_\ell$, so the cache-choice space is $2^{mp|T_\ell|}$ rather than $2^{mp}$).

We flag this gap explicitly here, rather than obscure it, to give the reader an accurate picture of which parts of the argument are rigorous and which depend on a yet-unverified probabilistic step.
\end{remark}

\begin{lemma}[Adversarial permutation with preserved trace]\label{lem:adversary}
Fix any algorithm with a locality-respecting controller. Fix a reference $\pi_1$ and exit index $t^\star \in (\tau_{j-1}, \tau_j]$ with $u := z_{t^\star - 1}(\pi_1) \notin T_1 := T(\pi_1)$, as provided by Lemma~\ref{lem:exit}. Define the target set
\[
  \mathcal{D} \;:=\; [n] \setminus \bigl(T_1 \cup \{z_0, \dots, z_k\}(\pi_1) \cup \pi_1(T_1)\bigr),
\]
of size $|\mathcal{D}| \geq n - 2(1+Ls) - (k+1)$. For each $d \in \mathcal{D}$, there is a permutation $\pi_d$ such that:
\begin{enumerate}[label=(\alph*), itemsep=1pt, topsep=1pt]
  \item $\pi_d|_{T_1} = \pi_1|_{T_1}$;
  \item by (a) and Lemma~\ref{lem:trace}, $\mathcal{T}(\pi_d) = \mathcal{T}(\pi_1)$, $T(\pi_d) = T_1$, and $X^{(\ell)}_0(\pi_d) = X^{(\ell)}_0(\pi_1)$ for all $\ell$;
  \item $z_{t^\star}(\pi_d) = d$, and distinct $d, d' \in \mathcal{D}$ yield distinct $z_{t^\star}$ (since $z_{t^\star}(\pi_d) = d$).
\end{enumerate}
\end{lemma}

\begin{proof}
\emph{Construction of $\pi_d$.} For $d \in \mathcal{D}$, observe $\pi_1^{-1}(d) \notin T_1$: if $\pi_1^{-1}(d) \in T_1$, then $d \in \pi_1(T_1)$, contradicting $d \notin \pi_1(T_1)$. Define
\[
  \pi_d(u) := d, \qquad \pi_d(\pi_1^{-1}(d)) := \pi_1(u), \qquad \pi_d(x) := \pi_1(x) \text{ for all } x \notin \{u, \pi_1^{-1}(d)\}.
\]
This is a value-swap (transposition in the codomain) of $\pi_1$, hence a permutation. Both modified positions lie in $[n] \setminus T_1$, so $\pi_d|_{T_1} = \pi_1|_{T_1}$, giving (a). Part (b) follows from (a) by Lemma~\ref{lem:trace}.

\emph{Part (c).} We track the chain of $\pi_d$ up to step $t^\star$. By Lemma~\ref{lem:exit}, there exists an exit step in window $j$, so (taking $\pi_1$ to realize this) the $\pi_1$-orbit of $z_0 = 1$ has length at least $t^\star$, i.e., $z_0, z_1, \dots, z_{t^\star - 1}$ are pairwise distinct under $\pi_1$. The chain up to $z_{t^\star - 1}$ is the same under $\pi_d$ as under $\pi_1$: this uses that $\pi_d$ agrees with $\pi_1$ on positions $z_0, z_1, \dots, z_{t^\star - 2}$. To see this: first note $z_0 = 1 \in T_1$. This is because a correct algorithm must compute $z_1 = \pi(1)$ as part of any multi-hop chain for $k \geq 1$; by Lemma~\ref{lem:reach}, $X^{(L)}_0$ depends on $\pi$ only through $\pi|_{T_L}$, so the state cannot depend on $\pi(1)$ unless $1 \in T_L$. Since the same argument applies at the first layer where the chain is consulted---and in particular whenever $\pi_1$ has a chain-in-trace prefix of positive length---$1 \in T_1$. Now, $\pi_d$ changes $\pi_1$ only at $u = z_{t^\star - 1}(\pi_1)$ and at $\pi_1^{-1}(d)$; neither of these positions lies on the $\pi_1$-chain prefix $z_0, \dots, z_{t^\star - 2}$. For $u = z_{t^\star - 1}$: excluded by definition of $t^\star$ (the first chain position outside $T_1$; all earlier $z_0, \dots, z_{t^\star - 2}$ either lie in $T_1$ or precede $u$ in the orbit, and $u$ itself is the $(t^\star - 1)$-th iterate, not an earlier one). For $\pi_1^{-1}(d)$: we have $d \notin \{z_1, \dots, z_k\}(\pi_1)$ by construction, which (together with $\pi_1$'s injectivity on the orbit) gives $\pi_1^{-1}(d) \notin \{z_0, \dots, z_{k-1}\}(\pi_1)$, in particular not among $z_0, \dots, z_{t^\star - 2}$. Hence $\pi_d|_{\{z_0, \dots, z_{t^\star - 2}\}} = \pi_1|_{\{z_0, \dots, z_{t^\star - 2}\}}$, and the $\pi_d$-chain from $z_0$ coincides with the $\pi_1$-chain up to $z_{t^\star - 1}$, with $z_{t^\star}(\pi_d) = \pi_d(z_{t^\star - 1}) = \pi_d(u) = d$ by construction.

At step $t^\star$: $z_{t^\star}(\pi_d) = \pi_d(z_{t^\star - 1}(\pi_d)) = \pi_d(u) = d$.

For distinct $d, d' \in \mathcal{D}$, we get distinct $z_{t^\star}$ values, as claimed.
\end{proof}

\begin{remark}[Stopping at $t^\star$ vs.\ reaching $\tau_j$]\label{rmk:t-star}
Lemma~\ref{lem:adversary} gives distinctness of $z_{t^\star}$, not directly $z_{\tau_j}$. When $t^\star = \tau_j$, this is what we want. When $t^\star < \tau_j$, the chain continues from $d$ for $\tau_j - t^\star$ more steps before reaching $z_{\tau_j}$, and \emph{a priori} two different $d, d' \in \mathcal{D}$ could produce the same $z_{\tau_j}$ through a chain collision (e.g., if $d$ lies in a $2$-cycle of $\pi_1$).

We avoid this in two steps, both developed in the proof of Lemma~\ref{lem:incompr}:
\begin{enumerate}[label=(\roman*), leftmargin=2em, itemsep=1pt]
  \item \emph{Fix the first-exit step $t^\star_0$ by pigeonhole.} The set $\mathcal{N}$ of permutations with a window-$j$ exit partitions by first-exit step into $\mathcal{N} = \bigsqcup_{t \in (\tau_{j-1}, \tau_j]} \mathcal{N}_t$ (at most $s$ parts). By pigeonhole at least one $\mathcal{N}_{t^\star_0}$ has density $\geq 1/s$ within $\mathcal{N}$. Restrict to $\pi_1 \in \mathcal{N}_{t^\star_0}$.
  \item \emph{Restrict further to $\pi_1$ with no short non-principal cycles.} For $s \leq \sqrt{n}/4$, permutations with all non-principal cycles of length $> s$ form a positive-density subset of the conditioned ensemble.
\end{enumerate}

Both restrictions reduce the set of \emph{valid reference permutations} we may choose, but once a single valid $\pi_1$ is fixed, the clean-target counting of Lemma~\ref{lem:incompr} gives distinct $z_{\tau_j}$ values across the entire clean set of targets---\emph{no $1/s$ loss propagates to the distinct-value count}. The $O(H \log s)$ gap between Conjecture~\ref{thm:lower} and Theorem~\ref{thm:upper} is thus attributable entirely to the upper bound's $\log_2(2s)$ factor (from windowed doubling's inner loop), not to the lower-bound proof.
\end{remark}

\begin{lemma}[Windows are sequential]\label{lem:sequential}
Assume $n \geq 4k$ and $s \leq \sqrt{n}/4$. Suppose the algorithm, at layer $\ell_j$, produces a representation $X^{(\ell_j)}_0$ that determines $z_{\tau_j}$. Then $\ell_j \cdot s \geq \tau_j = \min(js, k)$. In particular, $\ell_j \geq j$ when $js \leq k$. (Outside the regime $s \leq \sqrt{n}/4$, Proposition~\ref{prop:max-bound} unconditionally gives $\ell_j \cdot s \geq \tau_j - 1$; the unit-level strengthening here follows the proof strategy of Lemma~\ref{lem:incompr}.)
\end{lemma}

\begin{proof}
By Lemma~\ref{lem:reach}, $X^{(\ell_j)}_0$ depends only on inputs in $D_{\ell_j}(0) \subseteq T(\pi)$. Suppose for contradiction $\ell_j s < \tau_j$. Then $|T(\pi)| \leq 1 + \ell_j s < 1 + \tau_j$, so the chain $z_0, z_1, \dots, z_{\tau_j}$ contains at least one token $z_t \notin T(\pi)$.

Fix such a $\pi = \pi_1$ \emph{additionally satisfying the cycle-structure restriction of Lemma~\ref{lem:incompr}} (all non-principal cycles of $\pi_1$ have length $> s$). Such $\pi_1$ exist in the conditioned ensemble with positive density for $s \leq \sqrt{n}/4$. Take the smallest $t \in \{1, \dots, \tau_j\}$ with $z_{t-1}(\pi_1) \notin T_1 := T(\pi_1)$. Since $n \geq 4k$ and $|T_1| \leq 1 + \ell_j s \leq 1 + Ls$, the complement $\mathcal{D} := [n] \setminus (T_1 \cup \{z_0, \dots, z_k\} \cup \pi_1(T_1))$ has size $|\mathcal{D}| \geq n - 2(1 + Ls) - (k+1) \geq n/2 - k - O(1)$ in the regime $L \leq n/(8s)$ (otherwise $L \geq k/s$ already holds and the lemma is vacuous).

Apply Lemma~\ref{lem:adversary} to $\pi_1$ and target $d \in \mathcal{D}$: it yields $\pi_d$ with $\pi_d|_{T_1} = \pi_1|_{T_1}$, $\mathcal{T}(\pi_d) = \mathcal{T}(\pi_1)$, and by part~(c), $z_{t^\star}(\pi_d) = d$. Since $d \notin \{z_0, \dots, z_k\}(\pi_1)$, we have $z_{t^\star}(\pi_d) = d \neq z_{t^\star}(\pi_1)$. By Lemma~\ref{lem:trace}, $X^{(\ell_j)}_0(\pi_d) = X^{(\ell_j)}_0(\pi_1)$.

We now convert this distinctness at $t^\star$ into distinctness at $\tau_j$ for \emph{at least one} $d \in \mathcal{D}$. When $t^\star = \tau_j$: immediate. When $t^\star < \tau_j$: by the clean-target counting in the proof of Lemma~\ref{lem:incompr}, under the cycle-structure restriction on $\pi_1$, at most $s$ values of $d \in \mathcal{D}$ (``orbit hits $u$'' case) fail to satisfy $z_{\tau_j}(\pi_d) = \pi_1^{\tau_j - t^\star}(d)$; the ``orbit hits $\pi_1^{-1}(d)$'' case is excluded by the cycle restriction. Since $|\mathcal{D}| \geq n/2 - k - O(1) > s$ in the regime $n \geq 4k$ and $s \leq \sqrt{n}/4$, at least one clean $d \in \mathcal{D}$ exists; for it $z_{\tau_j}(\pi_d) = \pi_1^{\tau_j - t^\star}(d) \neq z_{\tau_j}(\pi_1)$ (by injectivity of $\pi_1^{\tau_j - t^\star}$ together with $d \neq z_{t^\star}(\pi_1)$). Combined with $X^{(\ell_j)}_0(\pi_d) = X^{(\ell_j)}_0(\pi_1)$, this contradicts the assumption. Hence $\ell_j s \geq \tau_j$.
\end{proof}

Lemma~\ref{lem:sequential} is only a statement about the \emph{earliest} layer at which window $j$'s output can appear. To get the lower bound on total depth, we also need to control the number of layers that window $j$ consumes \emph{after} window $j-1$'s output has been resolved. This is the content of Lemma~\ref{lem:incompr} and Lemma~\ref{lem:per-window} below.

\begin{lemma}[Window incompressibility]\label{lem:incompr}
Assume $n \geq 4k$ and that the cache controller is locality-respecting (Definition~\ref{def:local}). For any window index $j \in \{1, \dots, W\}$, there exist a layer $\ell^\dagger \leq L$ and a value $x^\dagger \in \mathcal{Q}_p^m$ such that, letting $\Pi^\dagger := \{\pi : X^{(\ell^\dagger)}_0(\pi) = x^\dagger\}$:
\begin{enumerate}[label=(\roman*), itemsep=1pt, topsep=1pt]
  \item for $j = 1$: $\ell^\dagger = 0$ and $x^\dagger = X^{(0)}_0$ is the fixed initial state (so $\Pi^\dagger = S_n$);
  \item for $j \geq 2$: $\ell^\dagger = \ell_{j-1}$ (the resolution layer of window $j-1$).
\end{enumerate}
In both cases, under the additional hypothesis $s \leq \sqrt{n}/4$, $z_{\tau_j}$ takes at least $\lfloor n/2 \rfloor - k - O(s)$ distinct values as $\pi$ ranges over $\Pi^\dagger$. In particular, this is $\geq n/4$ for $n \geq n_0$ sufficiently large.
\end{lemma}

\begin{proof}
By Lemma~\ref{lem:exit} (conditional on step $(\star)$, Remark~\ref{rmk:exit-rigor}), there exists a reference $\pi_1$ with window-$j$ chain exit. The exit index $t^\star \in (\tau_{j-1}, \tau_j]$ provided by Lemma~\ref{lem:exit} is not generally $\tau_j$, so we need an additional step to fix $t^\star$. By the pigeonhole argument of Remark~\ref{rmk:t-star}, among the permutations $\mathcal{N}$ satisfying the exit property, there exists at least one exit index $t^\star_0 \in (\tau_{j-1}, \tau_j]$ such that $|\mathcal{N}_{t^\star_0}|/|\mathcal{N}| \geq 1/s$; we restrict attention to such $\pi_1 \in \mathcal{N}_{t^\star_0}$. \emph{We do not assume $t^\star_0 = \tau_j$.}

Set $\ell^\dagger$ per cases (i)/(ii) and $x^\dagger := X^{(\ell^\dagger)}_0(\pi_1)$ (in case (i), $x^\dagger$ is the constant initial state, automatically).

Apply Lemma~\ref{lem:adversary} to this $\pi_1$ with exit index $t^\star = t^\star_0$. It produces a target set $\mathcal{D}$ of size $\geq n - 2(1+Ls) - (k+1) \geq n/2 - k - O(1)$. Lemma~\ref{lem:adversary}(c) gives $z_{t^\star_0}(\pi_d) = d$ with distinctness across $d \in \mathcal{D}$, and Lemma~\ref{lem:adversary}(b) gives $X^{(\ell^\dagger)}_0(\pi_d) = x^\dagger$, so $\pi_d \in \Pi^\dagger$.

The conclusion we want is distinctness of $z_{\tau_j}$, not of $z_{t^\star_0}$. When $t^\star_0 = \tau_j$ these coincide. When $t^\star_0 < \tau_j$, distinct $z_{t^\star_0}$ values could in principle produce coinciding $z_{\tau_j}$ via chain collision after the exit step. We now count collision-inducing targets under the hypothesis $s \leq \sqrt{n}/4$.

For $d \in \mathcal{D}$: Lemma~\ref{lem:adversary} constructs $\pi_d$ by swapping values of $\pi_1$ at $u := z_{t^\star_0 - 1}(\pi_1)$ and $w_d := \pi_1^{-1}(d)$ only. Write $\Delta := \tau_j - t^\star_0 \leq s - 1$. The chain of $\pi_d$ starting from $d = z_{t^\star_0}(\pi_d)$ follows $\pi_1$'s iterations from $d$ until it hits one of the two swapped positions. Call $d \in \mathcal{D}$ \emph{clean} if the $\pi_1$-orbit $d, \pi_1(d), \pi_1^2(d), \ldots, \pi_1^{\Delta-1}(d)$ is disjoint from $\{u, w_d\}$. For clean $d$, $\pi_d$'s chain from $d$ coincides with $\pi_1$'s chain from $d$ for the first $\Delta$ iterations, yielding $z_{\tau_j}(\pi_d) = \pi_1^{\Delta}(d)$, which is an injective function of $d$ since $\pi_1$ is a permutation.

We bound $|\mathcal{D} \setminus \text{clean}|$ by two contributions:
\begin{enumerate}[label=(\roman*),leftmargin=2em,itemsep=1pt]
\item \emph{Orbit hits $u$}: $\pi_1^r(d) = u$ for some $r \in [0, \Delta-1]$, i.e., $d \in \{\pi_1^{-r}(u) : r \in [0, \Delta-1]\}$, at most $\Delta \leq s$ values.
\item \emph{Orbit hits $w_d = \pi_1^{-1}(d)$}: $\pi_1^r(d) = \pi_1^{-1}(d)$ for some $r \in [0, \Delta-1]$, i.e., $\pi_1^{r+1}(d) = d$, i.e., $d$ lies in a $\pi_1$-cycle of length dividing $r+1 \in [1, \Delta]$. The number of such $d \in [n]$ is at most $\sum_{\ell \text{ divides some } r+1 \in [1, \Delta]} \ell \cdot c_\ell(\pi_1) \leq \sum_{\ell \leq \Delta} \ell \cdot c_\ell(\pi_1) \leq \Delta \cdot n \leq s n$ in worst case, where $c_\ell(\pi_1)$ is the number of $\pi_1$-cycles of length $\ell$. This worst-case bound is too weak.
\end{enumerate}
To control (ii), we restrict attention to reference permutations $\pi_1$ whose non-principal cycles all have length $> s$. For uniform $\pi_1 \in S_n$ conditioned on $\mathcal{D}_\text{chain}$ (cycle of $1$ has length $> k$), the probability of having no non-principal cycle of length $\leq s$ is a positive density $\rho_s > 0$ (not uniform in $s$: $\rho_s$ decreases as $s$ grows, but positivity for each fixed $s \leq \sqrt{n}/4$ is what we need). By classical cycle-structure asymptotics (e.g., Arratia--Tavar\'e), the number of cycles of each length $\ell$ is approximately Poisson with mean $1/\ell$ for $\ell \leq n^{1/2}$, so the probability of having zero non-principal cycles of length $\leq s$ satisfies $\rho_s \approx \prod_{\ell = 1}^{s} e^{-1/\ell} \geq (es)^{-1}$ for $s = o(\sqrt{n})$ (valid under our precondition $s \leq \sqrt{n}/4$). Thus a valid reference permutation $\pi_1$ (one satisfying both the exit condition of Lemma~\ref{lem:exit} and the cycle-structure restriction) exists so long as the first-exit-step pigeonhole density $1/s$ combined with $\rho_s \geq (es)^{-1}$ gives positive probability; this holds throughout the precondition regime. Once a valid $\pi_1$ is fixed, neither the first-exit-step pigeonhole nor the cycle-structure restriction propagates into the distinct-value count of $z_{\tau_j}$ (see final paragraph of proof and Remark~\ref{rmk:t-star}).

Under this restriction, contribution (ii) vanishes (no short non-principal cycles), so $|\mathcal{D} \setminus \text{clean}| \leq s$. Hence $|\text{clean}| \geq |\mathcal{D}| - s \geq n/2 - k - s$, and the injective map $d \mapsto \pi_1^{\Delta}(d)$ on clean $d$ gives $\geq n/2 - k - s$ distinct values of $z_{\tau_j}(\pi_d)$ on $\Pi^\dagger$. Note: the first-exit-step pigeonhole and cycle-structure restriction only reduce the \emph{number of valid reference permutations $\pi_1$}; once a single valid $\pi_1$ is fixed, the distinct-value count of $z_{\tau_j}$ over the resulting $\{\pi_d : d \in \text{clean}\} \subseteq \Pi^\dagger$ is $|$clean$|$ directly, with no $1/s$ factor. Thus the distinct-value count is $\geq n/2 - k - s \geq n/4$ in the regime $s \leq \sqrt{n}/4$ and $n \geq 4k$ for $n$ sufficiently large.

For $s > \sqrt{n}/4$, the cycle-restriction argument degrades and the product bound is not guaranteed by this proof strategy; we fall back to Proposition~\ref{prop:max-bound}, consistent with the regime condition of Conjecture~\ref{thm:lower}.
\end{proof}

\begin{remark}[Effect on Conjecture~\ref{thm:lower}'s constant]
The distinct-value count $n/2 - k - s \geq n/4$ gives a bandwidth factor of $\lceil \log_2(n/4)/(Hmp)\rceil = \lceil (\log_2 n - 2)/(Hmp)\rceil$, matching $\Omega(\log_2 n/(Hmp))$ up to an additive $O(1)$ in the numerator. No $\log s$ loss appears in the distinct-value count. (The $O(H\log s)$ gap between Conjecture~\ref{thm:lower} and Theorem~\ref{thm:upper} is therefore attributable to the upper bound's $\log_2(2s)$ factor from windowed doubling's inner loop, not to the lower-bound proof strategy.) In particular, Conjecture~\ref{thm:lower}'s statement $L = \Omega(\lceil k/s\rceil \cdot \lceil \log n/(Hmp)\rceil)$ holds (modulo $(\star)$) with a universal constant $c$ in the regime $s \leq \sqrt{n}/4$.
\end{remark}

\begin{lemma}[Per-window cost]\label{lem:per-window}
For $n \geq 4k$, $s \leq \sqrt{n}/4$, and a locality-respecting controller, any correct algorithm must satisfy $\ell_j - \ell_{j-1} \geq B - 1$ for every $j \in \{1, \dots, W\}$, where
\[
  B \;:=\; \left\lceil \frac{\log_2 n}{Hmp} \right\rceil
\]
and $\ell_0 := 0$. More precisely, $B' := \ell_j - \ell_{j-1} \geq \lceil \log_2(\lfloor n/2\rfloor - k - s)/(Hmp)\rceil \geq \lceil (\log_2 n - 2)/(Hmp)\rceil \geq B - 1$; the additive $-1$ is $O(1)$ and is absorbed into the universal constant. No $\log_2 s$ loss appears.
\end{lemma}

\begin{proof}
By Lemma~\ref{lem:incompr}, in the regime $s \leq \sqrt{n}/4$ there exist $\ell^\dagger$ and $x^\dagger$ such that $z_{\tau_j}$ takes at least $M := \lfloor n/2 \rfloor - k - s \geq n/4$ distinct values on $\Pi^\dagger := \{\pi : X^{(\ell^\dagger)}_0(\pi) = x^\dagger\}$, for $n \geq n_0$ sufficiently large. In case $j \geq 2$, $\ell^\dagger = \ell_{j-1}$; in case $j = 1$, $\ell^\dagger = 0 = \ell_0$.

A correct algorithm maps each distinct $z_{\tau_j}$ to a distinct final query-token state (otherwise two inputs with different correct answers would produce the same output), so in particular at layer $\ell_j$, the representation $X^{(\ell_j)}_0$ must take at least $M$ distinct values on $\Pi^\dagger$. By Lemma~\ref{lem:bandwidth} with starting state $x^\dagger$ and $B' := \ell_j - \ell^\dagger$ layers of evolution,
\[
  M \;\leq\; |\Sigma_{B'}(x^\dagger)| \;\leq\; 2^{B' Hmp},
\]
which gives $B' \geq \lceil \log_2 M / (Hmp) \rceil \geq \lceil (\log_2 n - 2)/(Hmp) \rceil \geq B - 1$ where $B := \lceil \log_2 n /(Hmp) \rceil$. Thus $\ell_j - \ell_{j-1} \geq B - 1$ (with $\ell_0 = 0$ for $j = 1$); the additive $-1$ is absorbed into the universal constant of Conjecture~\ref{thm:lower}.
\end{proof}

\begin{remark}[Dense regime $n = O(k)$]\label{rmk:dense}
The adversarial construction of Lemma~\ref{lem:incompr} requires $n \geq 4k$. In the dense regime $n = O(k)$, Proposition~\ref{prop:max-bound}'s unconditional bound $L \geq \max(\lceil (k-1)/s \rceil, \lceil \log_2 n/(Hmp)\rceil)$ still applies, giving $L = \Omega(\lceil k/s \rceil)$. The strengthening from $\lceil (k-1)/s \rceil$ to $\lceil k/s \rceil$ via Lemma~\ref{lem:sequential} applies in the same $s \leq \sqrt{n}/4$ regime as Conjecture~\ref{thm:lower} (since Lemma~\ref{lem:sequential}'s proof also invokes the clean-target counting of Lemma~\ref{lem:incompr}); in particular, this unit-level strengthening requires the cycle-structure restriction and is not claimed for general $s$.
\end{remark}

\begin{proof}[Proof strategy for Conjecture~\ref{thm:lower}, modulo $(\star)$]
Assuming the conclusion of Lemma~\ref{lem:exit} (exit exists at every window) and the regime $n \geq 4k$, $s \leq \sqrt{n}/4$, we combine Lemma~\ref{lem:sequential} and Lemma~\ref{lem:per-window}. Setting $\ell_0 := 0$, define $\ell_j$ as the earliest layer $\ell \in \{0, 1, \ldots, L\}$ such that $|T_\ell(\pi)| \geq \tau_j + 1$ for every $\pi$ in some non-empty set $\Pi^\dagger_j$ relevant to window $j$ (the set of $\pi$ satisfying $\mathcal{D}_{\text{chain}}$ and consistent with a fixed prefix state $x^\dagger$ at the resolution layer of window $j-1$, as in Lemma~\ref{lem:incompr}). This definition makes $\ell_j$ well-defined for every $j \leq W$: since $|T_L(\pi)| \geq \tau_j + 1$ for some $\pi \in \Pi^\dagger_j$ is necessary for a correct algorithm (by Lemma~\ref{lem:reach} applied to that $\pi$'s chain through $z_{\tau_j}$ to the final output $z_k$), the set of layers satisfying the condition is non-empty and contains $L$; its minimum is $\ell_j$. Moreover, $\ell_j$ is monotone in $j$ (since $\tau_j$ is), so $\ell_0 \leq \ell_1 \leq \cdots \leq \ell_W \leq L$. With this definition, Lemma~\ref{lem:sequential} gives $\ell_j \cdot s \geq \tau_j$ (rephrased: at layer $\ell_j$, $|T_{\ell_j}| \leq 1 + \ell_j s$ must be $\geq \tau_j + 1$, i.e., $\ell_j \cdot s \geq \tau_j$), and Lemma~\ref{lem:per-window} gives $\ell_j - \ell_{j-1} \geq B - 1$ where $B := \lceil \log_2 n/(Hmp)\rceil$.

\medskip

Now combine:

(a) \textbf{Sequentiality (Lemma~\ref{lem:sequential}):} $\ell_j \cdot s \geq \tau_j$ for each $j$; in particular $\ell_W \geq W$ when $\tau_W = k$.

(b) \textbf{Per-window cost (Lemma~\ref{lem:per-window}):} $\ell_j - \ell_{j-1} \geq B - 1$ for each $j \in \{1, \dots, W\}$, where $B := \lceil \log_2 n/(Hmp) \rceil$.

Summing (b) over $j = 1, \dots, W$ with $\ell_0 = 0$ gives
\[
  L \;\geq\; \ell_W \;=\; \sum_{j=1}^W (\ell_j - \ell_{j-1}) \;\geq\; W \cdot (B - 1) \;\geq\; \left\lceil \frac{k}{s} \right\rceil \cdot \left\lceil \frac{\log_2 n}{Hmp} \right\rceil \cdot \left(1 - \frac{1}{B}\right).
\]
For $n \geq n_0 := 100$ (which ensures $\sqrt{n}/4 \leq n/16$, hence $k + s \leq n/4 + n/16 \leq n/3$, hence $\lfloor n/2 \rfloor - k - s \geq n/6 \geq n_0/6 \geq 16$) and $\log_2 n > Hmp$ (so that $B \geq 2$; this is the condition for the bandwidth factor to be non-trivial---otherwise $\lceil \log_2 n/(Hmp)\rceil = 1$ and the product bound reduces to cache reachability, already covered by Proposition~\ref{prop:max-bound}), we have $1 - 1/B \geq 1/2$, yielding
\[
  L \;\geq\; \frac{1}{2} \cdot \left\lceil \frac{k}{s}\right\rceil \cdot \left\lceil \frac{\log_2 n}{Hmp} \right\rceil,
\]
i.e., Conjecture~\ref{thm:lower}'s statement with $c = 1/2$. For $s > \sqrt{n}/4$, the product-structure statement is not guaranteed by this proof strategy and we fall back to Proposition~\ref{prop:max-bound}'s max-bound, as stipulated in the Conjecture~\ref{thm:lower} precondition.

The derivation uses the conclusion of Lemma~\ref{lem:exit} (existence of exit at every window); since Lemma~\ref{lem:exit} is left as Open Problem~1, this establishes Conjecture~\ref{thm:lower} modulo $(\star)$ rather than unconditionally. No $\log_2 s$ loss appears in this derivation: the clean-target counting of Lemma~\ref{lem:incompr} reduces the set of valid reference permutations (via first-exit-step pigeonhole and cycle-structure restriction), but once a single valid $\pi_1$ is fixed, the distinct-value count of $z_{\tau_j}$ is $\Omega(n)$ directly.
\end{proof}

\paragraph{Why the product (not max) bound holds.}
The fact that each of the $W$ windows requires $B$ layers \emph{between resolutions}, not just total, is what gives the product $W \cdot B$ rather than $\max(W, B)$. Lemma~\ref{lem:per-window} is essential: window $j$'s resolution cannot be pipelined within the $B$ layers devoted to window $j-1$, because to distinguish among the $\Omega(n/s)$ possible values of $z_{\tau_j}$ requires $B$ \emph{fresh} layers past $\ell_{j-1}$.

\subsection{Remarks on Theorem~\ref{thm:barrier}}\label{sec:proof-barrier}

The proof of Theorem~\ref{thm:barrier} was given in Section~\ref{sec:main} following the theorem statement. Here we collect additional commentary on the barrier and the search for bypasses.

\begin{remark}[Breaking the barrier]\label{rmk:break-barrier}
Theorem~\ref{thm:barrier} does not rule out lower bounds outside the class of Definition~\ref{def:pwd}. Three natural avenues remain:
\begin{enumerate}[itemsep=1pt, topsep=1pt, leftmargin=1.5em]
  \item \emph{Unconditional communication-complexity lower bounds}: the pointer chasing problem has been studied in multi-party communication models. \citet{NW1993} proved a $\tilde\Omega(n/k^2)$ lower bound on the $k$-round randomized communication complexity of $k$-hop pointer chasing; \citep{Yehu2020} improved this to $\tilde\Omega(n/k - k \log n)$, essentially matching Klauck's earlier randomized lower bound and tight up to the $k\log n$ additive term relative to Nisan-Wigderson's $\tilde O(n/k)$ upper bound; \citep{Mao2025} closed the remaining $\log n$ gap to $\Omega(n/k + k)$ via gadgetless lifting. These CC bounds are \emph{unconditional}. What remains open is the \emph{lifting}: translating such a CC bound into an unconditional Cache-Transformer depth lower bound without invoking \citep{Sanford2024}'s MPC-conjecture reduction, which would bypass Definition~\ref{def:pwd}.
  \item \emph{Joint-state arguments}: counting over the joint state $(X^{(\ell)}_i)_{i \in S^{(\ell)}}$ rather than the single state $X^{(\ell)}_0$. The joint state has capacity $smp > Hmp$ whenever $s > H$, potentially escaping the per-layer bandwidth limit. However, in the $\PC_{n,k}$ task only $X^{(L)}_0$ decodes to the output, so arguments coupling to joint cache states require additional task structure (e.g., requiring multiple output tokens or global statistics).
  \item \emph{Information-theoretic multi-output tasks}: tasks whose output is $\Omega(Hmp)$ bits (e.g., requiring $\Omega(Hmp / \log n)$ chain values as simultaneous outputs at layer $L$) can give $L \geq \log(\text{output})/(Hmp) = \Omega(1)$ from information alone. Such tasks deviate from $\PC_{n,k}$ as stated, but may be natural models of multi-hop reasoning in different settings.
\end{enumerate}
Each of these is a distinct open problem.
\end{remark}

\begin{remark}[Exhaustive search for bypass arguments]\label{rmk:handles}
We considered eight alternative approaches to escape the barrier within counting-based arguments: (i) asymmetric rounds (parallel reasoning at multiple token positions); (ii) adversarial online streaming (cache commitment before seeing the rest of the input); (iii) query/cache capacity asymmetry ($smp$ vs $mp$); (iv) counting distinguishability at non-query tokens; (v) pebbling-game time-space tradeoffs; (vi) direct-sum and XOR lemmas for independent task instances; (vii) communication-complexity fooling-set arguments, in particular round-elimination lower bounds for pointer chasing of the \citet{DGS1987} style; and (viii) average-case (random-$\pi$) hardness. Each approach either saturates at $\lceil k/s\rceil$ or yields a strictly weaker bound (e.g., (vii) via unconditional MPC bounds gives $\log k / s$, weaker than $k/s$ for $k \geq 2$). We omit the detailed case-by-case analysis; the unified conclusion is the per-window structural limit formalized by Theorem~\ref{thm:barrier}.
\end{remark}

\subsection{Proof sketch of Theorem~\ref{thm:upper}}

\emph{Serial construction.} At each of $k$ outer layers, the query token performs a single content-based lookup: its current slot encodes the current $z_t$, keys encode token identities $i$, values encode payloads $\pi(i)$, and the attention output becomes the next $z_{t+1}$. Implementing one lookup requires $O(\lceil \log_2 n/(mp) \rceil)$ sub-layers to encode/decode $n$-ary pointers in $m$-dimensional $p$-bit vectors (standard construction, e.g., via binary decomposition). Total: $L = O(k \lceil \log_2 n/(mp) \rceil)$.

\emph{Windowed pointer doubling.} Partition the chain into $W = \lceil k/s \rceil$ windows of size $s$ each. Within each window, apply pointer doubling: at stage $t$, compute $\pi^{2^t}$ on the window's starting token using $\lceil \log_2 s \rceil$ stages, so window $j$ consumes $O(\lceil \log_2 s \rceil \cdot \lceil \log_2 n/(mp)\rceil)$ layers. Total: $L = O(\lceil k/s \rceil \cdot \lceil \log_2 s \rceil \cdot \lceil \log_2 n / (mp) \rceil)$. Take the minimum of the serial and windowed costs.

\subsection{Proofs of Theorems~\ref{thm:oblivious}, \ref{thm:adaptive}, and Corollary~\ref{cor:separation}}\label{sec:proof-coupling}

Fix $\pi \sim \mathrm{Unif}(S_n)$. Let $z_t := \pi^t(1)$ for $t \geq 0$, with $z_0 = 1$. The cache at stage $t$ is $C_t \subseteq [n]$, $|C_t| = s$. Success at stage $t$ is $\mathcal{E}_t := \{z_t \in C_t\}$, joint success is $\mathcal{E} := \cap_{t=1}^T \mathcal{E}_t$.

\subsubsection{Oblivious case: exponential decay (Theorem~\ref{thm:oblivious})}

The key idea is direct enumeration of $(z_1, \dots, z_T)$ trajectories rather than a chain-rule decomposition. Because $\pi$ is uniform, any fixed distinct trajectory has probability $(n-T)!/n!$; the joint hit event is a union over trajectories lying in $C_1 \times \cdots \times C_T$.

\begin{proof}[Proof of Theorem~\ref{thm:oblivious}]
Condition on the caches $C_1, \dots, C_T$ (oblivious, so independent of $\pi$). Define
\[
  A \;:=\; \bigl\{(a_1, \dots, a_T) \in C_1 \times \cdots \times C_T \;:\; a_1, \dots, a_T \text{ distinct}, \; a_t \neq 1 \text{ for all } t\bigr\}.
\]
Trivially $|A| \leq s^T$.

For any $(a_1, \dots, a_T) \in A$, $\Pr[z_1 = a_1, \dots, z_T = a_T]$ is the probability that $\pi(1) = a_1$, $\pi(a_1) = a_2$, \dots, $\pi(a_{T-1}) = a_T$. These are $T$ consistent constraints on the bijection $\pi$ (consistent because $1, a_1, \dots, a_T$ are distinct), giving probability $(n-T)!/n! = 1/(n(n-1)\cdots(n-T+1))$.

Let $\mathcal{E}^{\mathrm{good}} := \{z_1, \dots, z_T \text{ distinct, all} \neq 1\} \cap \mathcal{E}$ and $\mathcal{E}^{\mathrm{bad}} := \mathcal{E} \setminus \mathcal{E}^{\mathrm{good}}$. Then
\[
  \Pr[\mathcal{E}^{\mathrm{good}} \mid \{C_t\}] \;=\; \frac{|A|}{n(n-1)\cdots(n-T+1)} \;\leq\; \frac{s^T}{(n-T+1)^T} \;\leq\; \left(\frac{s}{n-T}\right)^T,
\]
using $n(n-1)\cdots(n-T+1) \geq (n-T+1)^T \geq (n-T)^T$ for $n \geq T + 1$.

For $\mathcal{E}^{\mathrm{bad}}$: this requires $z_i = z_j$ for some $i < j$, or $z_i = 1$ for some $i$. For the first, $z_i = z_j \iff \pi^{j-i}(z_i) = z_i$; by exchangeability of uniform permutations, $\Pr[\pi^m(y) = y]$ for any fixed $y$ equals the probability that $y$ lies in a cycle of length dividing $m$, which is at most $m/n$. Hence $\Pr[z_i = z_j] \leq (j-i)/n$. Summing with exact value: $\sum_{1 \leq i < j \leq T} (j-i) = \sum_{d=1}^{T-1} d(T-d) = T(T-1)(T+1)/6$, giving $\sum_{i<j} \Pr[z_i = z_j] \leq T(T^2-1)/(6n) \leq T^3/(6n)$. For the second, $\Pr[z_i = 1] = \Pr[\pi^i(1) = 1] \leq i/n$; summing $\sum_{i=1}^T i/n = T(T+1)/(2n) \leq T^2/n$. In total,
\[
  \Pr[\mathcal{E}^{\mathrm{bad}}] \;\leq\; \frac{T^3}{6n} + \frac{T^2}{n} \;\leq\; \frac{T^3 + 6T^2}{6n} \;\leq\; \frac{T^3}{3n} \qquad (T \geq 6),
\]
with a trivial explicit bound for $T \leq 5$. Using the uniform bound $\Pr[\mathcal{E}^{\mathrm{bad}}] \leq 2T^3/n$ (valid for all $T \geq 1$) for simplicity, we combine:
\[
  \Pr[\mathcal{E}] \;=\; \Pr[\mathcal{E}^{\mathrm{good}}] + \Pr[\mathcal{E}^{\mathrm{bad}}] \;\leq\; \left(\frac{s}{n-T}\right)^T + \frac{2T^3}{n}.
\]
(A sharper wrapper constant $T^3/(3n) + $ lower-order replaces $2T^3/n$ for $T \geq 6$, but does not change the asymptotic character of the bound.)
Taking expectation over $\{C_t\}$ (which has no $\pi$-dependence) preserves the bound.
\end{proof}

\subsubsection{Adaptive case: upper bound (Theorem~\ref{thm:adaptive}, upper bound)}

The upper bound uses the fact that at stage 1, any locality-respecting controller's cache $C_1$ is independent of $\pi(1)$ (because $C_1$ reads only positions in $T_0 = \{0\}$, which is disjoint from $\pi$'s domain $[n]$).

\begin{proof}[Proof of Theorem~\ref{thm:adaptive}, upper bound]
At stage 1, the trace $T_0 = \{0\}$ (only the query token is accessible before any attention computation), so $\pi|_{T_0}$ is vacuous. Hence $C_1$ is a deterministic function of the empty string, i.e., constant. Therefore $C_1 \perp \pi$, and since $\mathcal{E} \subseteq \mathcal{E}_1 = \{\pi(1) \in C_1\}$,
\[
  \Pr[\mathcal{E}] \;\leq\; \Pr[\pi(1) \in C_1] \;=\; \frac{|C_1|}{n} \;\leq\; \frac{s}{n} \;\leq\; \frac{s}{n-T},
\]
using that $\pi(1)$ is uniform on $[n]$ under $\pi \sim \mathrm{Unif}(S_n)$, $|C_1| \leq s$, and $n \geq T+1$.
\end{proof}

\subsubsection{Adaptive case: matching achievability (Theorem~\ref{thm:adaptive}, lower bound)}

We exhibit an explicit locality-respecting adaptive controller whose joint success probability matches the upper bound $s/n$ of Theorem~\ref{thm:adaptive} with exact equality. The strategy is ``chain-tracking'': at stage 1, gamble $s-1$ random slots plus the query's source position $1$ to learn $\pi(1)$; after a hit, the subsequent chain is deterministically known to the controller.

\begin{proof}[Proof of Theorem~\ref{thm:adaptive}, lower bound]
Define the controller $\mathsf{A}^\star$ as follows:
\begin{itemize}[itemsep=1pt,topsep=1pt,leftmargin=1.5em]
  \item \textbf{Stage 1:} $C_1 = \{1, a_1, a_2, \dots, a_{s-1}\}$ for any fixed distinct $a_i \in [n] \setminus \{1\}$. Note $1 \in C_1$.
  \item \textbf{Stage $t \geq 2$:} If $\mathcal{E}_1, \dots, \mathcal{E}_{t-1}$ all occurred (stage $t-1$ cache $C_{t-1} \ni z_{t-1}$), then by induction $z_1, \dots, z_{t-1} \in T_{t-1}$, and controller has read $\pi$ at these positions. In particular, controller knows $z_t = \pi(z_{t-1})$. Set $C_t \ni z_t$ (other $s-1$ slots arbitrary).
  \item If any earlier $\mathcal{E}_i$ failed, set $C_t$ arbitrarily (the joint event $\mathcal{E}$ has already failed).
\end{itemize}

\textbf{Locality check:} At stage $t \geq 2$, $C_t$ depends only on $\pi|_{\{1, z_1, \dots, z_{t-1}\}} \subseteq \pi|_{T_{t-1}}$ (all indices in prior caches). Locality holds.

\textbf{Success probability:} Under $\mathsf{A}^\star$, if $\pi(1) \in C_1$ then stage 1 succeeds ($z_1 = \pi(1) \in C_1 = \mathcal{E}_1$), and the chain-tracking induction at subsequent stages deterministically maintains $z_t \in C_t$ for all $t \geq 2$: the controller has read $\pi$ at all previously-visited chain positions (including $z_{t-1}$), so it knows $z_t = \pi(z_{t-1})$ and places it in $C_t$. This covers the fixed-point case $\pi(1) = 1$ as well: if $\pi(1) = 1$, the controller reads this at stage 1, knows $z_t = 1$ for all $t$, and includes $1$ in every $C_t$, so $\mathcal{E}$ holds. Therefore
\[
  \Pr[\mathcal{E}] \;\geq\; \Pr[\pi(1) \in C_1] \;=\; \frac{|C_1|}{n} \;=\; \frac{s}{n}.
\]
This matches the upper bound $s/n$ of Theorem~\ref{thm:adaptive} exactly (both bounds are $s/n$), giving $\Pr[\mathcal{E}] = s/n$.
\end{proof}

\begin{proof}[Proof of Corollary~\ref{cor:separation}]
Adaptive achieves $\Pr[\mathcal{E}] \geq s/n$ (Theorem~\ref{thm:adaptive}, achievability), while oblivious gives $\Pr[\mathcal{E}] \leq (s/(n-T))^T + 2T^3/n$. In the regime $s^T \geq 2T^3 n^{T-1}$ (equivalently $(s/n)^T \geq 2T^3/n$), the main term dominates the error term: since $(s/(n-T))^T \geq (s/n)^T \geq 2T^3/n$, we have $(s/(n-T))^T + 2T^3/n \leq 2 \cdot (s/(n-T))^T$.

To relate $(s/(n-T))^T$ to $(s/n)^T$, note $s/(n-T) = (s/n) \cdot n/(n-T) = (s/n) \cdot (1 + T/(n-T))$, so $(s/(n-T))^T = (s/n)^T \cdot (1 + T/(n-T))^T$. Under the mild condition $T \leq \sqrt{n}/2$ (so, using $n \geq 4$ to get $n - T \geq n/2$, we have $(1 + T/(n-T))^T \leq e^{T^2/(n-T)} \leq e^{(n/4)/(n/2)} = e^{1/2} \leq 2$), we have $(s/(n-T))^T \leq 2 (s/n)^T$. Hence
\[
  \frac{s/n}{(s/(n-T))^T + 2T^3/n} \;\geq\; \frac{s/n}{4 \cdot (s/n)^T} \;=\; \frac{(n/s)^{T-1}}{4} \;=\; \Omega\!\left((n/s)^{T-1}\right).
\]
Outside this regime (when $2T^3/n$ dominates), the ratio is $\geq (s/n) \div (4T^3/n) = \Omega(s/T^3)$, which, while weaker than $(n/s)^{T-1}$, still separates the two asymptotically for $s \gg T^3$.
\end{proof}

\section{Architecture Comparison}

\begin{table}[h]
\centering
\begin{tabular}{lccc}
\toprule
Architecture & Full cache ($s=n$) & \multicolumn{2}{c}{Cache $s < n$ [this work]} \\
\cmidrule(lr){3-4}
& & Lower & Upper \\
\midrule
Full softmax attention & $O(\log k)^\dagger$ & $\Omega(k/s)$ & $O((k/s)\log s)$ \\
Linear attention (small $r$) & $\Theta(k)$ & \multicolumn{2}{c}{$\Theta(k)$} \\
RNN / Mamba & $\Theta(k)$ & \multicolumn{2}{c}{$\Theta(k)$} \\
\bottomrule
\end{tabular}
\caption{Depth for $k$-hop pointer chasing. Our $\Omega(k/s)$ lower bound (Proposition~\ref{prop:max-bound}) is unconditional and parameter-free; it is meaningful when $s = o(k)$, reducing to the trivial $\Omega(1)$ in the full-cache limit $s = n$. In the $s = n$ limit, the meaningful unconditional lower bound comes from the bandwidth factor $\Omega(\log_2 n/(Hmp))$ of Proposition~\ref{prop:max-bound}, which for $Hmp = O(\log n)$ recovers $\Omega(\log n) \geq \Omega(\log k)$. Upper bound $O((k/s)\log s)$ assumes $mp \geq \log n$ (standard LLM precision). $\dagger$Upper bound from \citep{Sanford2024}; matching $\Omega(\log k)$ lower bound requires the MPC conjecture. The multiplicative gap between our lower and upper bounds is $O(H\log s)$ in general, reducing to $O(\log s)$ when $H = 1$.}
\end{table}

Full attention's depth scales as $\Omega(k/s)$ to $O((k/s)\log s)$, with a gap of $O(\log s)$. Linear attention and RNNs are stuck at $\Theta(k)$ regardless of cache. Full attention's advantage depends critically on cache size.

\section{Experiments}

We validate our bounds via \emph{constructive verification}: we hardcode the serial pointer-chasing algorithm directly into Transformer weights and measure accuracy on random permutations. This eliminates optimization confounds entirely---success or failure reflects theoretical possibility, not SGD's ability to find a solution.

\subsection{Hardcoded Transformer (serial strategy)}

We implement the serial strategy from Theorem~\ref{thm:upper}'s proof as explicit Transformer weights on $\PC_{16,k}$. Each layer performs one content-based lookup. We sweep $k \in \{1,2,4,8,12\}$ and $L \in \{1,\dots,16\}$ with 5000 random permutations per configuration. Results (Figure~\ref{fig:main}a): the transition from random guess ($1/n = 6.25\%$) to $100\%$ is sharp at exactly $L = k$.

\begin{figure}[htbp]
  \centering
  \includegraphics[width=\textwidth]{fig_main.pdf}
  \caption{Experimental validation on $\PC_{16,k}$ (5000 random permutations per configuration, no training). (a) Hardcoded Transformer with oracle cache: sharp $0\%\to100\%$ transition at $L=k$. (b) Windowed PD vs.\ serial for $k=8$: cache reduces required depth from $L=8$ to $L=3$--$4$. (c) Random (oblivious) cache: exponential error amplification with $k$, consistent with Theorem~\ref{thm:oblivious}'s $(s/(n-T))^T$ decay rate (cf.\ Remark~\ref{rmk:empirical} for numerical comparison at $n=16$). (d) Minimum depth heatmap for windowed PD.}
  \label{fig:main}
\end{figure}

\subsection{Windowed pointer doubling vs.\ serial}

\begin{figure}[htbp]
  \centering
  \includegraphics[width=\textwidth]{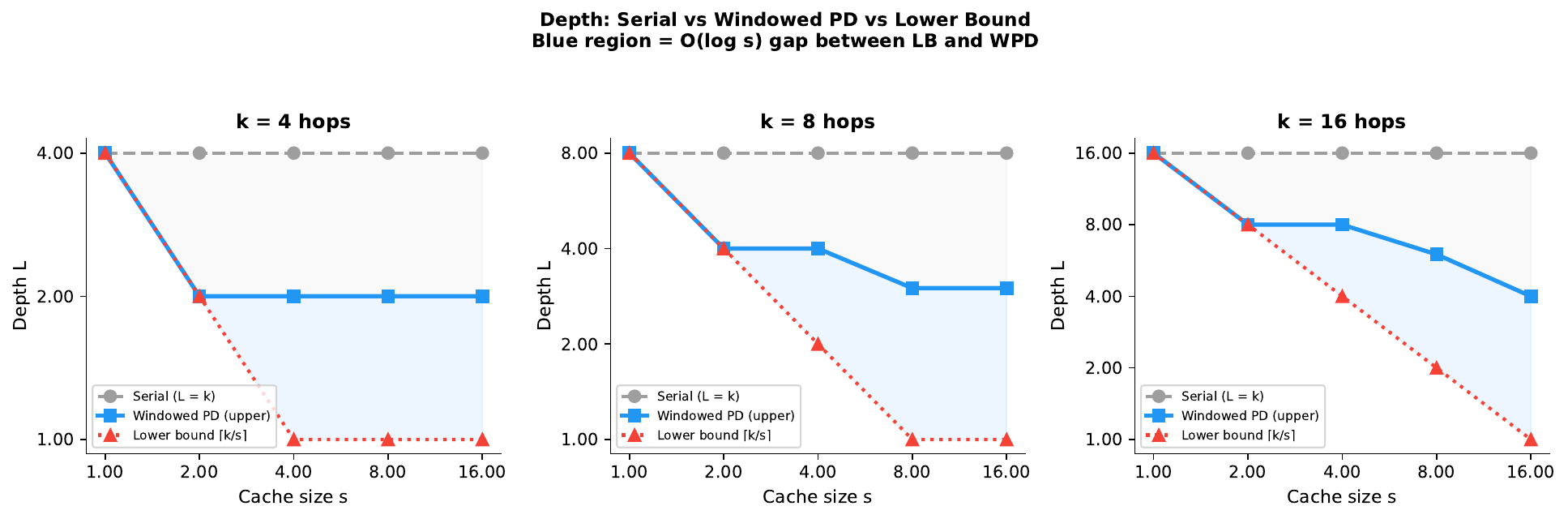}
  \caption{Depth--cache tradeoff for $k \in \{4,8,16\}$. Gray dashed: serial strategy ($L=k$, cache-independent). Blue: windowed pointer doubling upper bound. Red: lower bound $\lceil k/s\rceil$. Shaded blue region: the $O(\log s)$ gap between lower and upper bounds.}
  \label{fig:depth}
\end{figure}

We simulate oracle windowed pointer doubling on the same task, sweeping $s \in \{1,2,4,8,16\}$. Results (Figure~\ref{fig:main}b): for $k=8$, serial requires $L\geq 8$, windowed PD with $s=2$ achieves $100\%$ at $L=4$, and $s=8$ at $L=3$. Cache size trades for depth, as predicted. Figure~\ref{fig:depth} shows the full depth--cache tradeoff with the $O(\log s)$ gap between lower bound and windowed PD visible.

\subsection{Random cache: error amplification}

Figure~\ref{fig:main}c shows that under oblivious (random) cache eviction, joint success probability decays sharply with $k$. At $s = 8, n = 16, k = 8, T = 3$, we observe empirical success of $\approx 17.5\%$. The \emph{main term} of Theorem~\ref{thm:oblivious}'s upper bound is $(s/(n-T))^T = (8/13)^3 \approx 23.3\%$, which upper-bounds the good-chain success probability $\Pr[\mathcal{E}^{\mathrm{good}}]$ (not $\Pr[\mathcal{E}]$ in general; the empirical $17.5\%$ includes contributions from $\mathcal{E}^{\mathrm{bad}}$ not bounded by the main term alone). The main term matches the empirical value in order and confirms the qualitative $(s/(n-T))^T$ scaling. (The full theorem bound, including the $2T^3/n$ error term, exceeds 1 at this small $n$; see Remark~\ref{rmk:empirical}.) Note that the simpler expression $(s/n)^T = 12.5\%$ underestimates because it uses $n$ rather than $n-T$ in the denominator; Theorem~\ref{thm:oblivious} identifies the correct asymptotic decay rate as $(s/(n-T))^T$, which reduces to $(s/n)^T(1 + O(T^2/n))$ in the $s/n \to 0$ regime.

\subsection{Connection to LLM behavior}

\citet{Anan2026} independently report a sharp hallucination cliff near 90\% compression, with multi-hop (Hops) F1 dropping from ${\sim}31\%$ to near zero under aggressive compression, while single-hop retrieval degrades gracefully. This matches our framework qualitatively: $k$-hop capability is gated by a product bound that involves $k$, so multi-hop ($k \geq 4$) is hit harder than single-hop ($k=1$) as $s$ shrinks. \citet{Liu2025} show that adaptive (heavy-hitter) eviction outperforms recency- and norm-based baselines on reasoning; Corollary~\ref{cor:separation} gives a direct quantitative explanation: adaptive controllers achieve $\Theta(s/n)$ joint success, while oblivious ones decay as $(s/n)^T$, a separation factor of $\Omega((n/s)^{T-1})$. This is consistent with the observation that heavy-hitter-based methods, which concentrate cache on frequently-attended positions, provide qualitatively different error-scaling behavior than random or recency-based eviction. \emph{Caveat on regime.} Our experiments at $n = 16, T = 3$ do not fall within the ``tight'' regime of Corollary~\ref{cor:separation} (which requires $s \geq (2T^3)^{1/T} \cdot n^{(T-1)/T} \approx 24$, exceeding $n = 16$); we view the empirical observations here as qualitative validation of the direction of the separation rather than quantitative confirmation of the $(n/s)^{T-1}$ rate. Establishing the rate empirically in the main-term-dominates regime requires moderate-to-large $n$ (e.g., for $s/n = 1/2$ and $T = 3$, roughly $n > 432$; see Remark~\ref{rmk:empirical}).

\section{Related Work}

\paragraph{Transformer depth.}
\citet{Sanford2023} separated attention from RNNs via communication complexity. \citet{Sanford2024} proved Transformer--MPC equivalence. \citet{AlmanYu2025} gave SETH-conditional time bounds. \citet{Koza2025} proved infinite-precision lower bounds. We introduce cache as a new parameter. Our finite-precision model is analogous to bounded-precision circuit models in computational complexity; the key distinction from \citep{Sanford2024} is that their $\Omega(\log k)$ holds at infinite precision (conditional on MPC), while our conjectured product bound addresses the finite-precision regime (under a locality-respecting controller, which is satisfied by all standard KV-compression methods). The two results are complementary: neither implies the other.

\paragraph{KV cache compression.}
Eviction: \citep{Zhang2023,Liu2023,Li2024,Cai2024}. Reasoning-specific: \citep{Cai2025,Du2025}. Empirical: \citep{Liu2025,Anan2026}. All heuristic/empirical; we provide the first formal depth--cache tradeoff for a stylized reasoning task.

Recent methods employ question-aware or position-aware scoring (pseudo-query-based selection, global per-head budget optimization, head-aware risk-adaptive gating). In our framework, such strategies correspond to smarter choices of the cache set $S^{(\ell)}$ within the locality-respecting class; they may improve constant factors in practice but are subject to the unconditional max-bound $L = \Omega(\max(\lceil k/s\rceil,\lceil \log_2 n/(Hmp)\rceil))$ (Proposition~\ref{prop:max-bound}), and, if Conjecture~\ref{thm:lower} holds, cannot circumvent the product structure $L = \Omega(\lceil k/s\rceil \cdot \lceil \log_2 n/(Hmp)\rceil)$. The locality assumption is satisfied by all practical methods. Systems that offload KV entries to slower storage and retrieve on demand effectively increase $s$ per layer (at latency cost), which our bounds correctly capture. Soft compression (vector quantization, value merging) reduces effective precision or merges tokens rather than evicting them; extending our counting argument to this setting is an open direction.

\section{Discussion}\label{sec:disc}

We conjectured a product lower bound $L = \Omega((k/s) \cdot \log n/(Hmp))$ (Conjecture~\ref{thm:lower}) and isolated the sole remaining technical gap as the probabilistic step $(\star)$ in Remark~\ref{rmk:exit-rigor}. We unconditionally proved the max-bound $L = \Omega(\max(\lceil k/s\rceil,\lceil \log_2 n/(Hmp)\rceil))$ (Proposition~\ref{prop:max-bound}), and an unconditional matching upper bound $L = O(\lceil k/s\rceil \cdot \lceil \log_2(2s)\rceil \cdot \lceil \log_2 n/(mp)\rceil)$ via windowed pointer doubling, so if Conjecture~\ref{thm:lower} holds then the characterization is tight up to $O(H\log s)$ in the narrow-model regime. We also proved unconditionally a bandwidth barrier showing that the product bound cannot be sharpened in the $Hmp \geq \log_2 n$ regime by any per-window distinguishability-counting argument, and established that adaptive cache selection does not beat the oblivious bound at the first stage via a coupling argument. Theorems~\ref{thm:oblivious} and~\ref{thm:adaptive} (oblivious-vs-adaptive error behavior) are unconditional.

\paragraph{What the bandwidth barrier means.}
Theorem~\ref{thm:barrier} is the technically most significant structural statement of the paper. It explains a puzzling feature of existing cache-compression theory: why, despite decades of work on pointer chasing in communication complexity, we cannot get a depth lower bound for Transformers better than the trivial cache-reachability bound in the regime of practical interest. The barrier is that per-window output has $\log_2 n$ bits of ``capacity'' (since $z_{\tau_j} \in [n]$), while per-layer bandwidth is $Hmp$ bits; the ratio ceilings at $1$ whenever $Hmp \geq \log_2 n$. Breaking through requires either (i) output-informational arguments that escape the per-window output dimension (perhaps via multi-output tasks), (ii) translating the \emph{unconditional} pointer-chasing CC lower bounds of \citep{NW1993,Yehu2020,Mao2025} into Cache-Transformer depth lower bounds without the MPC reduction (an open lifting problem), or (iii) joint-state counting that couples to cache-wide state capacity $smp > Hmp$ when $s > H$---but this in turn requires the task to force $s$ cache tokens to jointly carry information.

\paragraph{Limitations and scope.}
Our model assumes hard cache boundaries, $p$-bit precision, and a locality-respecting cache controller. The precision assumption is essential for Lemma~\ref{lem:bandwidth} and reflects hardware reality. The locality assumption, which rules out controllers that examine representations outside the cache, is satisfied by all practical methods but is a genuine restriction relative to the most general adaptive controller. Our results thus complement \citep{Sanford2024}'s infinite-precision, MPC-conditional $\Omega(\log k)$, which is cache-oblivious but holds under unrestricted computation. When $s = n$ (full cache), our bound gives only $L \geq \lceil \log n/(Hmp)\rceil$, weaker than $\Omega(\log k)$ but holding unconditionally on the controller.

In standard LLM configurations (large $Hmp$), the bandwidth factor equals 1, and our bound reduces to $L \geq \lceil k/s\rceil$ alone---which holds unconditionally (without locality) via Lemma~\ref{lem:reach}. The full product structure is a \emph{structural} insight---cache and bandwidth are multiplicative, not competing---that becomes quantitatively binding in narrow or low-precision models.

Pointer chasing is a worst-case model; real reasoning tasks have exploitable structure permitting more aggressive compression than our bounds suggest. Our experiments use $n=16$ for clarity; the hardcoded construction's correctness is exact (independent of $n$ for $n < 2^p$), so the sharp threshold behavior transfers to any $n$ within the precision range.

\paragraph{Concurrent work.}
\citet{Anan2026} (arXiv March 2026) is a concurrent and independent empirical study. Their phase-transition-like behavior in multi-hop accuracy under compression is qualitatively consistent with our theory but was discovered independently.

\paragraph{Open problems.}
\begin{enumerate}[itemsep=2pt,topsep=2pt]
  \item \textbf{Close the trace--chain joint-distribution gap in Lemma~\ref{lem:exit}.} The probabilistic step $(\star)$ in Remark~\ref{rmk:exit-rigor}---bounding $\Pr[z_{\sigma_j}(\pi) \in T(\pi) \mid \mathcal{D}_{\text{chain}}] \leq \mathbb{E}[|T|]/(n-1)$, where $\sigma_j := \tau_j - 1$ is the chain position one step before window $j$'s boundary---is left unproven for adaptive, label-sensitive algorithms under uniform $\pi$. Two natural approaches: (a) a second-moment analysis using that the cycle structure of uniform $\pi$ decouples the orbit of $1$ from other positions in a controlled way, or (b) a Yao-principle reduction against an explicit adversarial distribution on $\pi$ that directly implies the existence of a window-$j$ exit. Closing this single step would turn Conjecture~\ref{thm:lower} into a theorem (in the $s \leq \sqrt{n}/4$ regime).
  \item \textbf{Extend the product bound to the large-cache regime $s > \sqrt{n}/4$.} Our proof strategy for Conjecture~\ref{thm:lower} relies on restricting the reference permutation's cycle structure (proof of Lemma~\ref{lem:incompr}), which degrades for $s > \sqrt{n}/4$ and forces us to state the conjecture only in the complementary regime. In the large-cache regime we establish only the unconditional max-bound of Proposition~\ref{prop:max-bound}. It is open whether the product structure $\lceil k/s \rceil \cdot \lceil \log_2 n/(Hmp)\rceil$ continues to hold for $s > \sqrt{n}/4$, or whether a different lower bound (e.g., a smooth interpolation between product and max in this range) is the truth.
  \item \textbf{Break the bandwidth barrier.} Theorem~\ref{thm:barrier} shows that any per-window distinguishability-based argument saturates at $\lceil k/s\rceil$ when $Hmp \geq \log_2 n$. Three avenues remain open: (i) translating the unconditional pointer-chasing CC lower bounds of \citep{NW1993,Yehu2020,Mao2025} into Cache-Transformer depth lower bounds without the MPC reduction, (ii) joint-state counting arguments coupling to cache-wide state capacity $smp$ rather than single-state $mp$, (iii) information-theoretic lower bounds on multi-output reasoning tasks that require $\Omega(Hmp)$ bits per layer of coupled output.
  \item \textbf{Remove the locality restriction.} Prove or disprove the product bound for fully adversarial cache controllers. Our adversarial construction (Lemma~\ref{lem:adversary}) uses locality to guarantee that perturbations outside $T_1$ are invisible to the controller; one possible path is to show the controller can ``leak'' at most $\log_2 \binom{n}{s}$ bits per layer via its $S^{(\ell)}$ choice, and to absorb this leakage into the bandwidth factor.
  \item Close the $O(H\log s)$ gap between lower and upper bounds.
  \item Extend the counting argument to soft compression (codebook size as effective bandwidth) and to infinite-precision models.
  \item Extend Theorems~\ref{thm:oblivious} and~\ref{thm:adaptive} beyond the uniform-random $\pi$ assumption to structured or adversarial inputs, which is closer to the empirical setting where pointer chains are structured (e.g., by entity relations in language data). Our oblivious bound's $(s/(n-T))^T$ decay rate depends on the uniformity of $\pi$; under structured $\pi$, both the oblivious decay rate and the adaptive-vs-oblivious separation may change.
  \item Validate on NLP multi-hop benchmarks with trained Transformers.
  \item Determine whether the product structure extends to reasoning tasks beyond pointer chasing, e.g., Boolean formula evaluation, tree-structured lookups, or graph reachability.
\end{enumerate}

\end{document}